%% file: main.tex
\documentclass{article}

\PassOptionsToPackage{numbers, compress}{natbib}  %
\usepackage{graphicx}

\usepackage[preprint]{neurips_2026}

\usepackage[utf8]{inputenc} %
\usepackage[T1]{fontenc}    %
\usepackage{hyperref}       %
\usepackage{url}            %
\usepackage{booktabs}       %
\usepackage{amsfonts}       %
\usepackage{nicefrac}       %
\usepackage{microtype}      %
\usepackage[table]{xcolor}  %
\usepackage{wrapfig}
\definecolor{mered}{RGB}{65, 105, 225}   %

\usepackage{amsmath}
\usepackage{colortbl}
\usepackage{todonotes}
\usepackage{siunitx}
\usepackage[most]{tcolorbox}
\newtcolorbox{promptbox}[1]{
  title=#1,
  colback=gray!5,
  colframe=gray!50,
  colbacktitle=gray!25,
  coltitle=black,
  fonttitle=\small\bfseries\ttfamily,
  boxrule=0.4pt,
  arc=2pt,
  left=8pt,
  right=8pt,
  top=4pt,
  bottom=4pt,
}
\title{How to Instruct Your Robot: Dense Language Annotations Power Robot Policy Learning}

\author{%
  Bosung Kim\textsuperscript{1,2}\thanks{Equal contribution.} \quad
  Ruiyi Wang\textsuperscript{1}\footnotemark[1] \quad
  David Acuna\textsuperscript{2} \quad
  Jaehun Jung\textsuperscript{2} \\
  \textbf{Alexander Trevithick}\textsuperscript{2} \quad
  \textbf{Brandon Cui}\textsuperscript{2} \quad
  \textbf{Yejin Choi}\textsuperscript{2} \quad
  \textbf{Prithviraj Ammanabrolu}\textsuperscript{1,2} \\
  \textsuperscript{1}University of California, San Diego \quad
  \textsuperscript{2}NVIDIA
}

\begin{document}

\maketitle

\begin{abstract}
Scaling robot policy learning is bottlenecked by the cost of collecting demonstrations: datasets at modern scale require thousands of skilled-operator hours on dedicated robot hardware.
The language description paired with these demonstrations does not face the same bottleneck---a single short task label is cheap, but it also leaves implicit spatial relations, object interactions, embodiment state, and subgoal structure that the pixels already contain. 
We treat \emph{language density} as a cheap lever for amplifying signal in a fixed demonstration corpus; whereas prior dense-language work in robotics commits to a single caption style, we instead ask which kind of dense language helps each task and learn to deliver it at deployment. We realize this as \textbf{DeMiAn} (\textbf{De}nse \textbf{M}ult\textbf{i}-aspect \textbf{A}nnotatio\textbf{n}) in two stages.
First, an automatic VLM pipeline re-labels each segment of an existing demonstration along four complementary aspects---\emph{physical motion}, \emph{scene composition}, \emph{arm pose}, and segment-level \emph{reasoning}---each surfacing a distinct kind of structure that a one-line task label omits.
Second, a small learned \emph{instructor}, trained via supervised fine-tuning, maps the natural language task description and an initial scene snapshot to a task-appropriate annotation and runs asynchronously alongside the action policy, hiding generation latency behind the rollout.
Applied to more than 1M robot manipulation and 50K EgoVerse human-egocentric videos, with no new demonstrations collected, DeMiAn delivers four findings on a VLA action policy and a video-based world-action model: 
(i)~the learned instructor raises RoboCasa success rate by 5 percentage points over the no-annotation baseline, within 3 points of a per-task oracle; (ii)~that oracle is non-trivial---no fixed aspect dominates, and peak performance requires selecting the right annotation aspect per task; (iii)~the trained system extends usefully to composite tasks under subgoal-driven prompt switching, and to OOD scenes and objects; and (iv)~dense annotation improves the compute-performance frontier in both mid-training and post-training, making re-annotation a practical scaling lever for robot policy learning.
\end{abstract}

\input{sections/intro}
\input{sections/related_work}
\input{sections/method}

\input{sections/experiments}

\input{sections/conclusion}

{
\small
\bibliographystyle{plainnat}
\bibliography{reference}
}

\newpage
\appendix
\input{sections/appendix}

\end{document}

%% file: sections/intro.tex
\begin{figure}[htbp]
    \centering
    \includegraphics[width=1\linewidth]{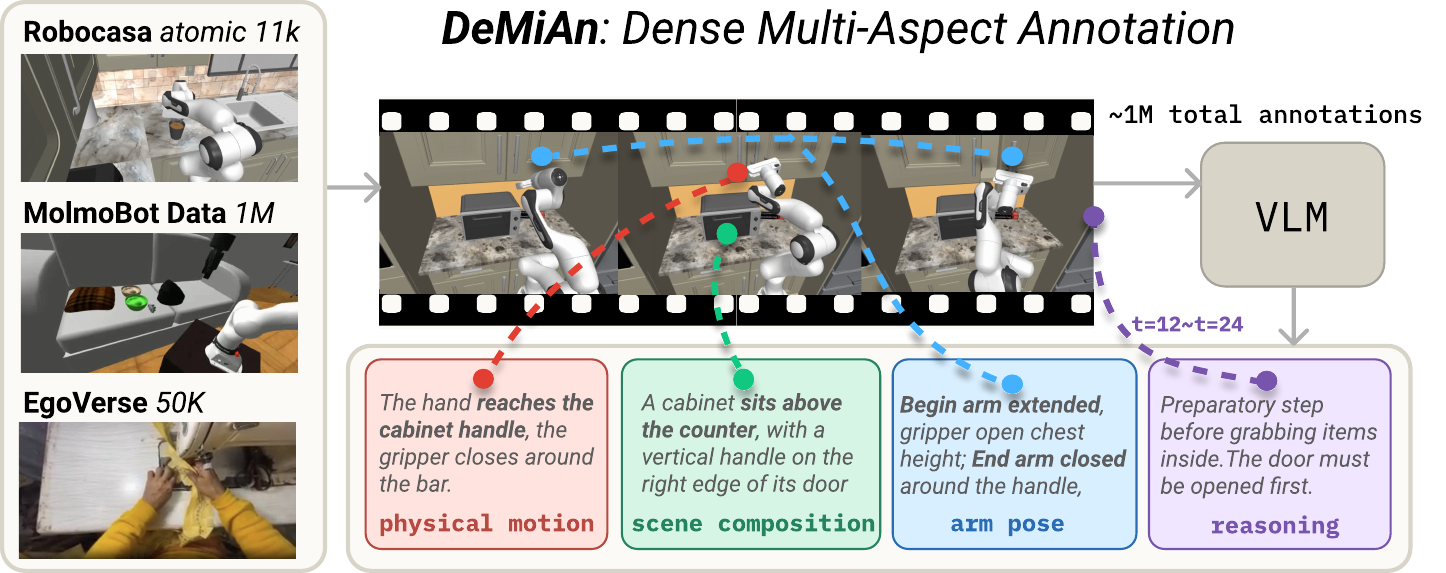}
    \caption{\textbf{Overview of DeMiAn.} We re-annotate existing robot and human demonstrations along four aspects: \emph{physical motion}, \emph{scene composition}, \emph{arm pose}, and \emph{reasoning}. We apply DeMiAn to 11K RoboCasa 365 clips, a 1M-scale MolmoBot dataset, and 50K EgoVerse human-egocentric clips.}
    \label{fig:teaser}
\end{figure}

\section{Introduction}

The dominant paradigm in robot learning today is scaling: more demonstrations, more environments, more diverse embodiments~\citep{walke2023bridgedata,open_x_embodiment_rt_x_2023,khazatsky2025droidlargescaleinthewildrobot}.
Vision-language-action (VLA) models have emerged at the center of this effort, unifying visual perception, language understanding, and action generation within a single architecture~\citep{Brohan2022RT1RT,pmlr-v229-zitkovich23a,black2026pi0visionlanguageactionflowmodel,kim2024openvla,pmlr-v305-black25a}. 
A more recent shift toward video-based world-action models (WAMs) extends this trajectory by mid-training on human-egocentric or web video before any robot data is seen, leveraging the fact that web video datasets scale far beyond robot teleoperation~\citep{nvidia2025dreamdojo,gear2025dreamzero,nvidia2025cosmos_predict}. 
Across both directions, the bottleneck is the same: datasets at modern scale require thousands of skilled-operator hours on dedicated robot hardware before any policy training begins, and even human-egocentric datasets demand similarly heavy capture and curation---EgoVerse 50K~\citep{egoverse}, for example, was assembled over $\sim$1{,}500 hours of in-person recording across many participants before any policy training could begin.

The language paired with these demonstrations does not face the same bottleneck. 
A typical clip is labeled with a short instruction such as ``open the drawer'' that names the goal but leaves implicit the spatial relations, contact transitions, embodiment state, and subgoal structure already present in the pixels and recorded actions.
Re-annotating an existing dataset with richer descriptions costs orders of magnitude less than collecting new demonstrations: a single VLM call produces a multi-sentence caption in seconds at sub-cent cost and with no skilled human labor, while every new demonstration requires teleoperation time, dedicated hardware, and environment setup. 
We treat \emph{language density} as a cheap lever for extracting more signal from a fixed demonstration dataset.

We instantiate this lever as \textbf{DeMiAn} (\textbf{De}nse \textbf{M}ult\textbf{i}-aspect \textbf{A}nnotatio\textbf{n}), an end-to-end approach with two stages: a training-time pipeline that re-annotates existing demonstration videos along multiple aspects, and an inference-time \emph{instructor} that supplies a task-appropriate aspect at deployment (\S\ref{sec:method}). 
Whereas prior dense-language work in robotics commits to a single style of caption---motion primitives~\citep{belkhale2024rth}, step-level chain-of-thought~\citep{zawalski2024ecot}, or captions used for representation pretraining~\citep{nair2022r3m,karamcheti2023voltron}---and uses it as a fixed training signal, we treat which aspect to use as a per-task design choice and learn to make that choice at deployment.
The pipeline produces four aspects per segment---\emph{physical motion}, \emph{scene composition}, \emph{arm pose}, and segment-level \emph{reasoning}---each surfacing a distinct kind of structure that a one-line task label omits, and we apply it to 11K RoboCasa 365 clips~\citep{robocasa,robocasa365}, a 1M-scale MolmoBot dataset~\citep{deshpande2026molmob0tlargescalesimulationenables}, and 50K EgoVerse human-egocentric clips~\citep{egoverse} without collecting any new demonstrations.
Different aspects help different tasks (\S\ref{sec:rq1-heterogeneity}), so any fixed-aspect deployment would underperform an oracle that picks per task; the instructor learns---via supervised fine-tuning---to approximate the per-task oracle from an initial scene snapshot and runs asynchronously at no added latency, turning a fixed demonstration dataset into a per-task source of language signal at deployment.

We evaluate DeMiAn on two popular robot policy architectures: \emph{DeMiAn VLA}, built on \texttt{openpi}~0.5~\citep{black2026pi0visionlanguageactionflowmodel}, and \emph{DeMiAn WAM}, a video-based world-action model~\citep{nvidia2025cosmos_predict}. We first show that a small learned instructor closes most of the per-task oracle gap from a single scene snapshot, raising RoboCasa success rate by 5 percentage points over the no-annotation baseline and reaching within 3 points of the oracle (\S\ref{sec:rq2-routing}). The trained system extends to composite tasks and shows stronger OOD generalization on MolmoSpaces scenes and objects than the task-only baseline (\S\ref{sec:rq3-generalization}). We further examine the compute-performance trade-off and find that dense annotation improves the frontier in both VLA post-training and WAM mid-training even after annotation-generation FLOPs are charged into the budget, making re-annotation a practical compute-efficient lever (\S\ref{sec:rq4-scaling}). 

Taken together, our contributions are: (i)~an automatic VLM re-annotation pipeline that surfaces four aspects of structure latent in existing robot and human demonstrations; (ii)~a learned \emph{instructor}, trained via supervised fine-tuning and deployed asynchronously at no added latency, that closes most of the per-task oracle gap from a single scene snapshot; (iii)~empirical evidence that this combination generalizes beyond the atomic training distribution to composite tasks and OOD scenes; and (iv)~a controlled compute-matched comparison showing that re-annotation improves the compute-performance frontier in both VLA post-training and WAM mid-training, even when annotation-generation FLOPs are charged into the budget.
At 1M-clip MolmoBot scale, DeMiAn matches the no-annotation baseline on MolmoSpaces NextTo and Color with $\sim$62\% less compute, positioning dense re-annotation as a complementary, compute-efficient lever alongside demonstration scaling.

%% file: sections/related_work.tex
\section{Related Work}
\subsection{Language-Conditioned Robotic Manipulation}
\textbf{Task-instruction conditioning.}
A substantial body of work conditions robot policies on a single short language instruction. CLIPort~\citep{shridhar2022cliport} grounds instructions as pixel-wise affordances via CLIP~\citep{radford2021clip} representations; BC-Z~\citep{jang2022bcz} conditions imitation learning on task descriptions for zero-shot generalization; VIMA~\citep{jiang2023vima} extends this to interleaved multimodal prompts. More recent vision-language-action (VLA) models scale this recipe dramatically: PaLM-E~\citep{driess2023palme} injects embodied observations into a pretrained language model, while RT-2~\citep{brohan2023rt2} and OpenVLA~\citep{kim2024openvla} co-finetune vision-language backbones on internet-scale data and robot trajectories, producing policies that inherit broad semantic knowledge from pretraining.

\textbf{Sub-instruction language.}
A few recent efforts introduce language at a finer temporal grain: RT-H~\citep{belkhale2024rth} inserts ``language motions'' between tasks and actions, embodied chain-of-thought approaches~\citep{zawalski2024ecot} emit step-level reasoning before actions, and \citet{ahn2022icanisay} grounds LLM plans in learned affordances.

In most of these policies, language serves primarily as a \emph{task identifier} (e.g., ``pick up the apple''), with generalization evaluated by varying the instruction while holding the description style fixed; finer-grained variants like RT-H~\citep{belkhale2024rth} and ECoT~\citep{zawalski2024ecot} relax this but each commits to a single fixed style of sub-instruction. Our work instead asks \emph{what kind} of dense language best supports policy learning---pairing each training frame with one of four aspect-specific captions---and finds that the choice of caption type matters as much as its presence.

\vspace{-0.5em}
\subsection{Dense Language Supervision and Reasoning for Embodied Agents}
\textbf{Inference-time language reasoning.}
A growing body of work leverages language-based reasoning in embodied settings at \emph{inference time}.
Chain-of-thought prompting~\citep{wei2022cot} established that eliciting step-by-step reasoning substantially improves LLM performance on complex tasks.
Inner Monologue~\citep{huang2023inner} closes the loop around LLM planners by feeding environment language feedback back into the prompt.
Code as Policies~\citep{liang2023code} and Language Models as Zero-Shot Planners~\citep{huang2022language} instead have LLMs emit executable programs or action sequences that invoke low-level robot skills.
ProgPrompt~\citep{singh2023progprompt} generates situated plans by combining LLM program synthesis with environment state assertions.
Across this line of work, language reasoning guides high-level decision-making at test time; the low-level policy itself receives only a short task instruction.

\textbf{Training-time language supervision.}
A complementary line of work uses language as a training signal rather than an inference-time aid.
R3M~\citep{nair2022r3m} and Voltron~\citep{karamcheti2023voltron} pretrain visual representations by aligning video frames with captions of human activity, producing features that transfer to downstream manipulation, and LIV~\citep{ma2023liv} unifies language-image value learning and reward specification within a single pretraining objective.
ROSIE~\citep{yu2023rosie} instead treats language as a data-augmentation interface, using text-to-image models to synthesize novel scenes from prompts and expand the visual diversity of robot training data.
Thought Cloning~\citep{hu2023thoughtcloning} trains agents to generate natural-language thoughts while acting, using human think-aloud demonstrations as supervision.

Unlike prior work that uses language for representation pretraining or data augmentation, we provide dense per-step captions as an auxiliary co-training signal during VLA fine-tuning, with captions required only at training time. We further systematically vary the \emph{type} of language supervision and find that the optimal caption type differs across manipulation primitives.

%% file: sections/method.tex
\section{DeMiAn: Dense Multi-Aspect Annotation}
\label{sec:method}
DeMiAn amplifies the language signal in existing demonstrations without collecting new data through two stages: a training-time \emph{annotation pipeline} that re-labels each video segment along multiple complementary aspects, and an inference-time \emph{instructor} that selects an appropriate aspect per task at deployment. We describe each in turn below.

\textbf{Choice of annotation aspects.}
Most language-supervised robot policies commit to a single sub-instruction style: motion primitives in RT-H~\citep{belkhale2024rth}, step-by-step chain-of-thought rationales in ECoT~\citep{zawalski2024ecot}, scene and object grounding in CLIPort~\citep{shridhar2022cliport}, or environment-state monologue in Inner Monologue~\citep{huang2023inner}. 
Each style has been shown to help, but along a different axis of the manipulation problem. 
We consolidate these paradigms into four complementary annotation \emph{aspects}---each a distinct dimension of the interaction described by its own caption---chosen to cover the structural information that a one-line task label leaves implicit (Table~\ref{tab:caption_types}): \texttt{physical\_motion} for temporal action and contact grounding, \texttt{scene\_composition} for spatial grounding, \texttt{arm\_pose} for embodiment and contact-state grounding, and \texttt{reasoning} for causal and subgoal grounding. 
Each segment is independently annotated along all four axes, so a downstream model can be trained on any single aspect, on a mixture, or with the aspect varying per segment.

\begin{table}[htbp]
\centering
\small
\caption{The four dense-annotation aspects used by DeMiAn. Each aspect targets and enriches information that is initially weakly specified by a one-line task description.}
\begin{tabular}{@{}l>{\raggedright\arraybackslash}p{3.2cm}>{\raggedright\arraybackslash}p{7.0cm}@{}}
\toprule
\textbf{Annotation aspect} & \textbf{Motivation} & \textbf{Annotation signal} \\
\midrule
\texttt{physical\_motion} & Temporal action and contact grounding. & Hand or end-effector motion: reaching, grasping, lifting, placing, opening, closing, and the objects involved. \\
\hline
\texttt{scene\_composition} & Spatial grounding. & Workspace, visible objects, distractors, and task-relevant spatial relations. \\
\hline
\texttt{arm\_pose} & Embodiment and contact-state grounding. & Hand or arm posture, reach direction, and contact state at segment boundaries. \\
\hline
\texttt{reasoning} & Causal and subgoal grounding. & Segment purpose (preparation, main manipulation, cleanup) and dependencies on neighboring segments. \\
\bottomrule
\end{tabular}
\label{tab:caption_types}
\end{table}

\textbf{Annotation pipeline.}
A \emph{segment} is a contiguous sub-interval of an episode that the dataset metadata associates with a single one-line label (e.g., a frame range labeled ``open the drawer''). For each segment, we sample up to $F{=}10$ uniformly spaced frames and issue one VLM call per aspect $k \in \mathcal{K} = \{\texttt{physical\_motion}, \texttt{scene\_composition}, \texttt{arm\_pose}, \texttt{reasoning}\}$. Each call is conditioned on the sampled frames and a context block containing the task description, scene descriptor, object list, and neighboring segment labels; prompts enforce a strict JSON schema and a two-sentence length cap. The output is one caption per aspect for the segment.

We apply this pipeline to three existing datasets without collecting new
demonstrations: RoboCasa~365~\citep{robocasa,robocasa365},
MolmoBot~\citep{deshpande2026molmob0tlargescalesimulationenables} (1M clips), and
EgoVerse~50K~\citep{egoverse}. RoboCasa atomic tasks can span multiple primitive
motions, so we split long demonstrations into single-primitive clips before
annotation (e.g., \texttt{OpenDoubleDoor} becomes two single-door clips); MolmoBot
clips are short enough to annotate as-is. For EgoVerse, we recast the four aspects
in human-egocentric language (hand motion and object transitions, egocentric
workspace, observable hand/arm state, subgoal links); see Appendix~\ref{sec:annotation-prompts}
for the adapted prompts. The labeling model is
Qwen3-VL-30B-A3B-Instruct~\citep{bai2025qwen3vltechnicalreport}.

\begin{wrapfigure}{r}{0.5\linewidth}
    \vspace{-10pt}
    \centering
    \includegraphics[width=0.95\linewidth]{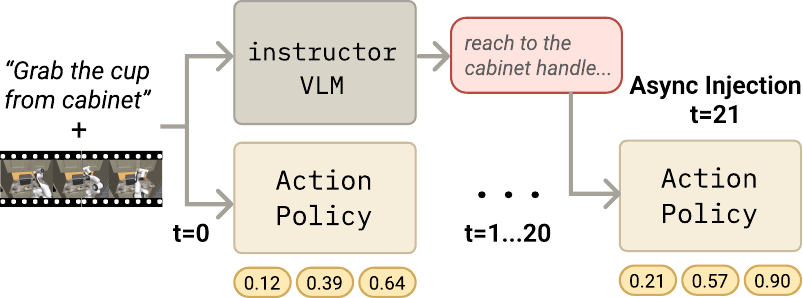}
    \caption{Asynchronous Instruction Injection.}             
    \label{fig:async}                                  
    \vspace{-10pt}
  \end{wrapfigure}
  
\textbf{Learned Instructor.}
The instructor is trained via supervised fine-tuning with reward-weighted target sampling. We first construct a reward table $w(\tau, k)$ by running the action policy with each of the four GT fixed-aspect annotations across all training tasks and recording per-task validation SR. For each training episode, the target aspect is sampled from a softmax over $w(\tau, \cdot)$ (temperature $T{=}2$, top-3 truncation), and the matching pipeline-generated caption is used as the SFT target. Tasks where every aspect underperforms the no-annotation baseline are assigned an empty target, teaching the instructor to abstain. Full details in Appendix~\ref{sec:routing-results}.

At deployment, the generated annotation is concatenated with the task description and placed in the same prompt slot. The instructor runs \textit{asynchronously} alongside the action policy: the robot begins acting with the task description alone, and the generated annotation is inserted at the next action-chunk boundary once it is ready (Figure~\ref{fig:async}). This hides instruction-generation latency behind the rollout and preserves the action server's normal control cadence.

%% file: sections/experiments.tex
\section{Experiments}
\label{sec:exp}

Our experiments evaluate how DeMiAn improves and scales robot policy learning along four axes. 
We characterize how each annotation aspect affects per-task performance and quantify the per-task oracle gap (\S\ref{sec:rq1-heterogeneity}).
We evaluate the trained instructor against that oracle (\S\ref{sec:rq2-routing}). 
We test generalization beyond the atomic training distribution to composite-task chaining and OOD scenes and objects (\S\ref{sec:rq3-generalization}). 
We characterize how DeMiAn scales in WAM mid-training and VLA post-training relative to growing the demonstration dataset (\S\ref{sec:rq4-scaling}).

\begin{figure}[htbp]
    \centering
    \includegraphics[width=1.0\linewidth]{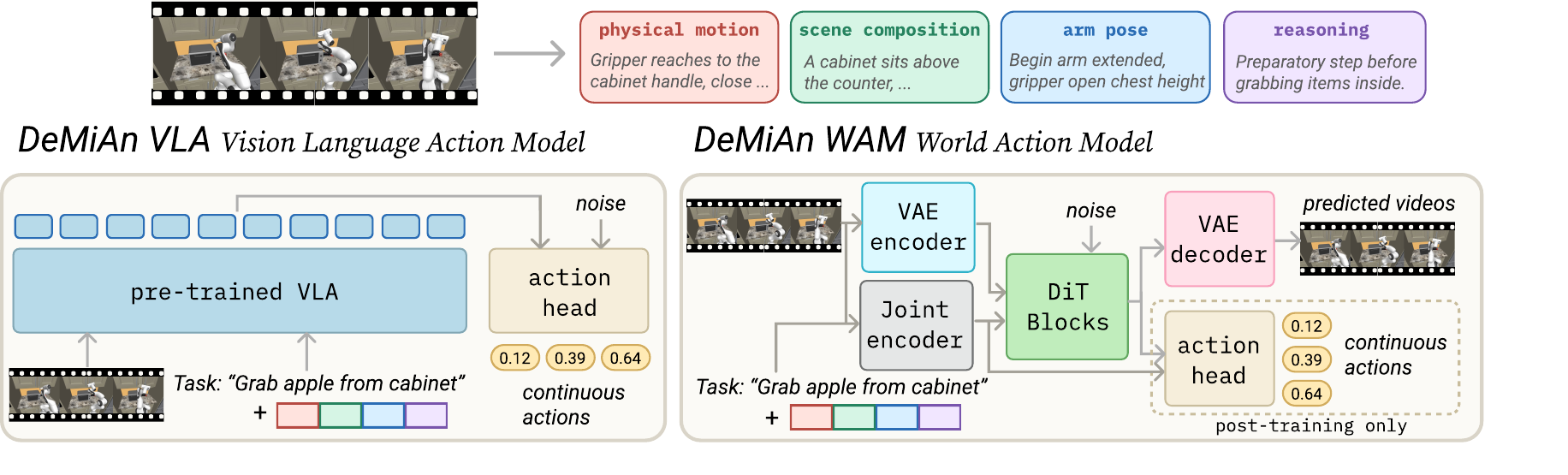}
    \caption{\textbf{Overview of the policy-model architectures} (DeMiAn VLA and DeMiAn WAM).}
    \label{fig:demian_train}
\end{figure}

\textbf{Policy models.}
Our experiments cover two representative robot policy architectures: vision-language-action policy (VLA) and world-model-based action policy (WAM), which span the mainstream paradigms in modern robot policy learning.
VLAs inherit broad semantic priors from large-scale image-text pre-training; WAMs instead build their backbones from raw videos at scales far beyond robot teleoperation, enabling transfer across embodiments. 
Testing DeMiAn against both lets us investigate whether the language density lever is architecture-specific or generalizes across this language-conditioned / video-pretrained divide.
We call the two architectures \textit{DeMiAn VLA} and \textit{DeMiAn WAM}, respectively, with details shown in Figure~\ref{fig:demian_train}.
\textbf{DeMiAn VLA:} We use the open-source \texttt{openpi}~0.5~\citep{black2026pi0visionlanguageactionflowmodel} as the VLA backbone, a PaliGemma vision-language backbone paired with a flow-matching action expert.
Relative to the original \texttt{openpi}, we add a small annotation-level LM cross-entropy ($\lambda_{\text{LM}}=0.1$) as a regularizer rather than a generative objective, which prevents the bidirectional decoder from compressing the annotation into a coarse global summary. 
\textbf{DeMiAn WAM:} We follow prior work that adapts video-generative world models for robot control by decoding actions from their learned spatiotemporal representations~\citep{wu2023unleashinglargescalevideogenerative,kim2026cosmospolicyfinetuningvideo,gear2025dreamzero} with a Cosmos-Predict 2.5 DiT backbone~\citep{nvidia2025cosmos_predict}.
Unlike joint video-action denoising, the video objective is unchanged and actions are predicted deterministically from the learned video features.
Across all annotation ablations the model architecture, visual data, and action targets are held fixed; only the language paired with each demonstration changes.
Full training objective and the rationale are in Appendix~\ref{sec:training-details}.

\textbf{Evaluation protocol.}
We use two benchmark suites for evaluation: RoboCasa365 and MolmoSpaces.
RoboCasa365 evaluation uses 17 atomic kitchen tasks in the held-out \texttt{target} split.
MolmoSpaces bench-v2~\citep{kim2026molmospaces} comprises nine benchmarks across four task families (Pick, Pick+Place (P+P), Pick+Place+NextTo, Pick+Place+Color) and varies difficulty along arm-initialization noise and scene distribution.
We report success rate (SR) throughout.
For RoboCasa365, given the task goal, an episode is successful if the simulator's task checker returns success during the rollout.
For MolmoSpaces, SR is evaluated on deterministic benchmark episodes and counts an episode successful if the task-specific success condition holds continuously for at least 0.5s at any point during the rollout.
Episode counts, seeds, and full protocol details are in Appendix~\ref{sec:eval-protocol-detail}.

\definecolor{avgpastel}{HTML}{FFF59D}
\begin{table}[t]
\centering
\caption{Success rate for training-time annotation ablations on RoboCasa365 and MolmoSpaces, grouped by policy architecture. Values are fractions in $[0,1]$; higher is better. \underline{Underlined} values mark the per-column best among the five training-time annotation types (Baseline through Reasoning); \textbf{bold} values mark the per-column best across all rows including \emph{w/ instructor}. The \emph{Per-task Best (oracle)} row is excluded from this comparison. MolmoSpaces columns summarize Table~\ref{tab:language-study-molmospace-detail}: Pick and P+P average the Standard (Std) and Hard splits, NextTo uses In-Distribution (ID), Color is unchanged, and Avg. (in \textcolor{blue}{blue}) averages these four family summaries.}
\label{tab:language-study}
\setlength{\tabcolsep}{2pt}
\scriptsize
\resizebox{\textwidth}{!}{%
\begin{tabular}{@{}l *{17}{c} c c c c c c@{}}
\toprule
 & \multicolumn{18}{c}{RoboCasa365} & \multicolumn{5}{c}{MolmoSpaces} \\
\cmidrule(lr){2-19} \cmidrule(lr){20-24}
Annotation aspect
 & \rotatebox{90}{CloseBlenderLid}
 & \rotatebox{90}{CloseFridge}
 & \rotatebox{90}{CloseToasterOvenDoor}
 & \rotatebox{90}{CoffeeSetupMug}
 & \rotatebox{90}{OpenCabinet}
 & \rotatebox{90}{OpenDrawer}
 & \rotatebox{90}{OpenStandMixerHead}
 & \rotatebox{90}{PnPCounterToCabinet}
 & \rotatebox{90}{PnPCounterToStove}
 & \rotatebox{90}{PnPDrawerToCounter}
 & \rotatebox{90}{PnPSinkToCounter}
 & \rotatebox{90}{PnPToasterToCounter}
 & \rotatebox{90}{SlideDishwasherRack}
 & \rotatebox{90}{TurnOffStove}
 & \rotatebox{90}{TurnOnElectricKettle}
 & \rotatebox{90}{TurnOnMicrowave}
 & \rotatebox{90}{TurnOnSinkFaucet}
 & \rotatebox{90}{\textcolor{blue}{Avg.}}
 & \rotatebox{90}{Pick (Avg of Std/Hard)}
 & \rotatebox{90}{P+P (Avg of Std/Hard)}
 & \rotatebox{90}{NextTo}
 & \rotatebox{90}{Color}
 & \rotatebox{90}{\textcolor{blue}{Avg.}} \\
\midrule
\rowcolor{black!8}
\multicolumn{24}{@{}l}{\textbf{DeMiAn VLA}} \\
Baseline (task-only) & .02 & .65 & .60 & .29 & .35 & .10 & \textbf{\underline{.84}} & .53 & .50 & \underline{.20} & \underline{.76} & .55 & .38 & .05 & \textbf{\underline{.86}} & \underline{.42} & \textbf{\underline{.43}} & \textcolor{blue}{.44} & .48 & \textbf{\underline{.64}} & .25 & .41 & \textcolor{blue}{.44} \\
Physical Motion   & \textbf{\underline{.03}} & .63 & \textbf{\underline{.74}} & \textbf{\underline{.60}} & .38 & .12 & .76 & .61 & \underline{.58} & .15 & .75 & .59 & \textbf{\underline{.75}} & \textbf{\underline{.10}} & .73 & .18 & .18 & \textcolor{blue}{\underline{.46}} & .59 & .60 & .20 & \underline{.46} & \textcolor{blue}{.46} \\
Scene Composition & .02 & .64 & .52 & .46 & \textbf{\underline{.42}} & \underline{.14} & .77 & \textbf{\underline{.62}} & .57 & .10 & .73 & \underline{.62} & .65 & .08 & .73 & .26 & .17 & \textcolor{blue}{.44} & \textbf{\underline{.61}} & .60 & .26 & .40 & \textcolor{blue}{.47} \\
Arm Pose          & \textbf{\underline{.03}} & \textbf{\underline{.71}} & .66 & .32 & .38 & .06 & .73 & .42 & .50 & .15 & .69 & .33 & .46 & .05 & .67 & .07 & .15 & \textcolor{blue}{.38} & .54 & .60 & .18 & .45 & \textcolor{blue}{.44} \\
Reasoning         & \textbf{\underline{.03}} & .54 & .40 & .36 & .26 & .05 & .72 & .48 & .52 & .08 & .62 & .34 & .50 & .05 & .66 & .26 & .40 & \textcolor{blue}{.37} & .59 & .60 & \textbf{\underline{.33}} & .39 & \textcolor{blue}{\underline{.48}} \\
\midrule
\rowcolor{blue!8}
w/ instructor & \textbf{.03} & .47 & .55 & .53 & .41 & \textbf{.47} & .80 & .53 & \textbf{.64} & \textbf{.21} & \textbf{.76} & \textbf{.72} & .55 & \textbf{.10} & .72 & \textbf{.48} & \textbf{.43} & \textcolor{blue}{\textbf{.49}} & .60 & .60 & .27 & \textbf{.48} & \textcolor{blue}{\textbf{.49}} \\
Per-task Best (oracle)     & .03 & .71 & .74 & .60 & .42 & .14 & .84 & .62 & .58 & .20 & .76 & .62 & .75 & .10 & .86 & .42 & .43 & \textcolor{blue}{.52} & .61 & .64 & .33 & .46 & \textcolor{blue}{.51} \\
\midrule
\rowcolor{black!8}
\multicolumn{24}{@{}l}{\textbf{DeMiAn WAM}} \\
Baseline (task-only) & \textbf{\underline{.02}} & .26 & .16 & .16 & .12 & \underline{.06} & \textbf{\underline{.35}} & .23 & .22 & .04 & .26 & \underline{.25} & .27 & \textbf{\underline{.06}} & \underline{.30} & \underline{.13} & .14 & \textcolor{blue}{\underline{.18}} & .26 & .19 & .06 & .14 & \textcolor{blue}{.16} \\
Physical Motion   & .01 & .25 & \textbf{\underline{.30}} & \textbf{\underline{.24}} & .15 & .05 & .30 & .24 & \textbf{\underline{.23}} & \underline{.06} & \textbf{\underline{.30}} & .24 & \textbf{\underline{.30}} & .04 & .29 & .07 & .07 & \textcolor{blue}{\underline{.18}} & \underline{.30} & \underline{.21} & .07 & \underline{.15} & \textcolor{blue}{\underline{.18}} \\
Scene Composition & .01 & .26 & .21 & .18 & \underline{.17} & \underline{.06} & .31 & \underline{.25} & \textbf{\underline{.23}} & .04 & .29 & \underline{.25} & .26 & .03 & .29 & .10 & .07 & \textcolor{blue}{\underline{.18}} & \underline{.30} & .20 & .09 & .13 & \textcolor{blue}{\underline{.18}} \\
Arm Pose          & .01 & \textbf{\underline{.28}} & .26 & .13 & .15 & .02 & .29 & .17 & .20 & \underline{.06} & .28 & .13 & .18 & .02 & .27 & .03 & .06 & \textcolor{blue}{.15} & .25 & .17 & .06 & \underline{.15} & \textcolor{blue}{.16} \\
Reasoning         & .01 & .22 & .16 & .14 & .10 & .02 & .29 & .19 & .21 & .03 & .25 & .14 & .20 & .02 & .26 & .10 & \underline{.16} & \textcolor{blue}{.15} & .28 & .19 & \textbf{\underline{.11}} & .13 & \textcolor{blue}{\underline{.18}} \\
\midrule
\rowcolor{blue!8}
w/ instructor     & \textbf{.02} & .14 & .29 & .10 & \textbf{.19} & \textbf{.13} & .30 & \textbf{.26} & .22 & \textbf{.08} & .24 & \textbf{.26} & .29 & .05 & \textbf{.32} & \textbf{.18} & \textbf{.18} & \textcolor{blue}{\textbf{.19}} & \textbf{.31} & \textbf{.23} & .09 & \textbf{.16} & \textcolor{blue}{\textbf{.20}} \\
Per-task Best (oracle)     & .02 & .28 & .30 & .24 & .17 & .06 & .35 & .25 & .23 & .06 & .30 & .25 & .30 & .06 & .30 & .13 & .16 & \textcolor{blue}{.20} & .30 & .21 & .11 & .15 & \textcolor{blue}{.19} \\
\bottomrule
\end{tabular}%
}
\end{table}

\subsection{RQ1: Which annotation type works best for each task?}
\label{sec:rq1-heterogeneity}
\label{sec:language-study}

We train both action policies with each of the four annotation aspects independently and with the original task-only label (\emph{Baseline}), and measure RoboCasa365 and MolmoSpaces SR at fixed compute. We also report a \emph{Per-task Best (oracle)} row that takes the column-wise maximum across the annotation rows, an upper bound over fixed annotation aspects. 

\textbf{The oracle gap.}
Table~\ref{tab:language-study} shows that dense annotation helps, but not uniformly.
On RoboCasa with DeMiAn VLA, \emph{Physical Motion} delivers large per-task lifts over the task-only baseline:
\emph{SlideDishwasherRack} jumps from 38\% to 75\% with Physical Motion, \emph{CoffeeSetupMug} from 29\% to 60\%, and \emph{CloseToasterOvenDoor} from 60\% to 74\%, while \emph{Arm Pose} and \emph{Reasoning} are markedly less useful.
On MolmoSpaces, \emph{Scene Composition} adds 13 percentage points on \emph{Pick} tasks, and the \emph{Reasoning} aspect improves 8 percentage points over the baseline in the \emph{NextTo} task.
Since no fixed aspect wins everywhere, we ask: \emph{what if an oracle instructor could pick the right aspect for each task?}
The per-task oracle reaches {52\% Avg.\ SR}, an 8 percentage point lift over the baseline and 6 percentage points above the best fixed aspect (\emph{Physical Motion}, 46\%).
DeMiAn WAM shows the same shape at lower absolute SR (best fixed aspect 18\%, oracle 20\%).
However, the oracle is unrealizable at deployment, where the right aspect is not known a priori---closing this gap is the goal of our learned instructor approach (\S\ref{sec:rq2-routing}).

\begin{figure}[t]
\centering
\includegraphics[width=\linewidth]{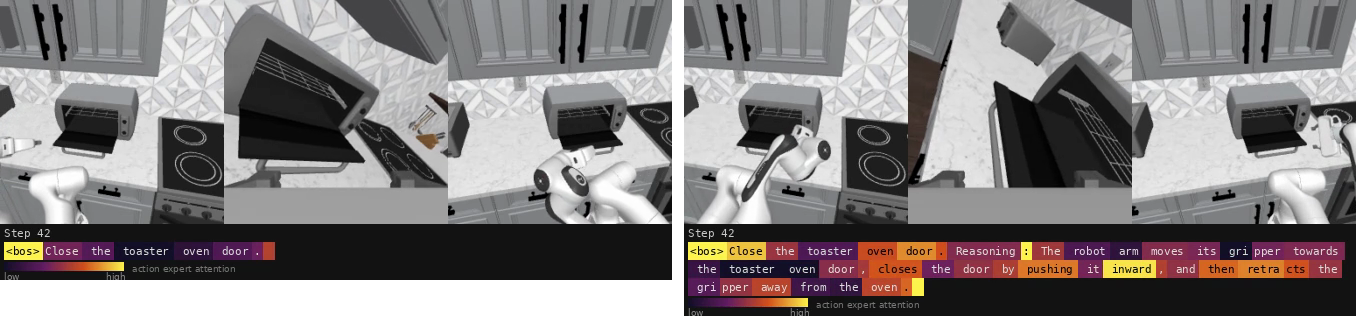}
\caption{Action expert attention over prefix tokens at step~42 of a \emph{CloseToasterOvenDoor} rollout.
\textbf{Left:} task-only baseline---attention is collapsed onto the \texttt{<bos>} token, with the task description ``Close the toaster oven door''.
\textbf{Right:} policy trained with \emph{Physical Motion} annotations---attention spreads across object mentions (\emph{oven door}), motion verbs (\emph{push}, \emph{retracts}), and directional adverbs (\emph{inward}, \emph{away}) in the annotation, indicating that the action expert is actively reading the dense caption rather than the task label alone.}
\label{fig:attn-close-toaster-oven-door}
\end{figure}

\textbf{What does the model learn from dense annotation?}
With the observation that some aspects can be dramatically helpful on specific tasks (e.g., 37 percentage points in \emph{SlideDishwasherRack} with \emph{Physical Motion}), we probe \emph{what} the action modules are reading from the prompt by visualizing their attention over prefix tokens during rollouts and inspecting the resulting patterns across tasks.
Figure~\ref{fig:attn-close-toaster-oven-door} shows an example on \emph{CloseToasterOvenDoor}. In the baseline policy, attention over prompt tokens is overwhelmingly concentrated on the \texttt{<bos>} token, with negligible weight on the task description that follows; the prompt is, in effect, being used as a position anchor rather than being read as language. In the Physical-Motion-annotated policy, attention instead distributes across linguistically grounded units of the annotation: interacting objects (\emph{``oven door''}), motion verbs not present in the task label (\emph{``push''}, \emph{``retracts''}), and directional adverbs (\emph{``inward''}, \emph{``away''}).

\textbf{Implication.}
No fixed annotation rule reaches the per-task oracle. Per-task and per-family breakdowns---which aspect wins on which task, where the baseline is strongest, and how the patterns differ on RoboCasa vs MolmoSpaces---are deferred to Appendix~\ref{sec:per-task-analysis}. The optimal aspect varies with the structural demands of each task family (contact-changing motion vs.\ open-vocabulary grounding); closing this gap requires producing the right instruction for each task at deployment, which is precisely what RQ2 evaluates.

\subsection{RQ2: Can a learned instructor approach the per-task oracle?}
\label{sec:rq2-routing}
Our analysis shows that no fixed annotation rule reaches the per-task oracle, and that the optimal aspect varies with the structural demands of each task family.
We close this gap with a small learned \emph{instructor} model: given the same observation the action policy receives, it generates a scene-grounded annotation.
We use Qwen3.5-2B~\citep{qwenteam2026qwen35omnitechnicalreport} as the instructor model.
To make the action policy robust to whichever annotation the instructor emits at inference, we train it on prompts sampled uniformly across five conditions: the four DeMiAn aspects and the task-only baseline.
At inference the instructor consumes (i)~three RGB camera frames from the initial observation of the rollout and (ii)~the natural-language task description. The output is concatenated to the task description and provided to the action policy as the conditioning prompt, occupying the same prompt slot used in \S~\ref{sec:rq1-heterogeneity}.

\textbf{Instructor closes the oracle gap.} The instructor raises average SR from 44\% to 49\% on RoboCasa and MolmoSpaces (Table~\ref{tab:language-study}), within 2--3 points of the per-task oracle (51--52\%).
To isolate the instructor's contribution, we compare against a random per-episode aspect baseline (46.6\% from Table~\ref{tab:async-instructor}) with the same action policy checkpoint: the instructor adds +3.8 points, indicating that the gain comes from learned per-task selection rather than heuristic routing alone.
Comparisons with fixed-aspect GT injection are in Appendix~\ref{sec:routing-results};
per-task and per-family analyses are in Appendix~\ref{sec:per-task-analysis}.

\textbf{Async deployment introduces no policy delay.}
We test async deployment for real-time applications. At test time the instructor decodes asynchronously while actions emit at the action server's native 8-step open-loop cadence. Until the instruction is ready, the policy executes on the task description alone, equivalent to the no-annotation baseline; the moment the instruction completes it is spliced into the reasoning field at the next chunk boundary, and subsequent action chunks condition on it.

We compare this design against a \emph{synchronous} variant that blocks the first action until the instruction completes (Table~\ref{tab:async-vs-sync}). Async tracks sync within fractional points on SR (49.0\% vs 49.5\%) while injecting the instruction into a rollout already in progress: the policy never waits for the instructor, and the $\sim$ 1.87 s of generation latency is hidden behind the first $\sim$ 22 action steps.

\begin{table}[h]
\centering
\caption{Async vs.\ sync instruction delivery comparison. \emph{Mean injected step} is the action step at which the instruction first reaches the policy. \emph{Mean injected time} is the mean wall-clock generation latency of the instructor for the same requests; in sync mode this latency directly delays the policy, whereas in async mode it overlaps with action execution and is therefore not added to rollout wall-clock.}
\label{tab:async-vs-sync}
\small
\begin{tabular}{lccc}
\toprule
Mode & Success Rate & Mean/Median injected step & Mean/Median injected time (s) \\
\midrule
Sync Instructor                & 49.5\%           & 1                & 0 \\
Async Instructor & 49.0\%   & 21.9/23               & 1.87/1.86 \\
\bottomrule
\end{tabular}
\end{table}

\subsection{RQ3: Does the method generalize on new instructions beyond the training distribution?}
\label{sec:rq3-generalization}

Our policy models are trained on the RoboCasa365 atomic tasks.
We test generalization to new instruction formats on composite tasks. We curate 18 composite tasks, each composed of atomic tasks seen during training, so chaining the atomic-task policies should in principle suffice.
We evaluate two prompting strategies to see how the action policy generalizes \emph{unseen instructions}. \emph{-fix} feeds a single composite task description for the whole episode; \emph{-dynamic} runs a subgoal-driven state machine that swaps the prompt to the in-distribution atomic instruction for the current phase once a lenient in-simulator trigger fires.
This yields four base configurations (\texttt{baseline-fix}, \texttt{baseline-dynamic}, \texttt{ours-fix}, \texttt{ours-dynamic}) where GT instructions are available at test time; we additionally evaluate \texttt{ours-dynamic-instructor}, which replaces the GT atomic prompt with the DeMiAn instructor's output and tests the full method end-to-end.

Table~\ref{tab:fix-vs-dynamic-example} illustrates the difference on \emph{PrepareCoffee}: \texttt{-fix} feeds a single composite description across the entire episode, while \texttt{-dynamic} swaps the prompt to the in-distribution atomic instruction for the current phase.
Each (task, config) is evaluated over 20 episodes with \texttt{max\_steps}=1200. We report phase-$k$ SR (fraction of episodes whose subgoal checker reaches the $k$-th milestone) and full-task SR via RoboCasa's strict success-goal condition checker.

\begin{table}[htbp]
\centering
\caption{Example of prompting for RoboCasa365 Composite task \emph{PrepareCoffee} (3-phase composite: pick mug $\to$ place under dispenser $\to$ press start). \texttt{-fix} provides one composite-task description for the entire episode. \texttt{-dynamic} swaps to the atomic instruction for the current phase, triggered when the policy's done-flag exceeds threshold and a lenient in-simulator condition fires.}
\label{tab:fix-vs-dynamic-example}
\small
\definecolor{phasefix}{HTML}{F2E6F8}      %
\definecolor{phaseone}{HTML}{E8F4FB}      %
\definecolor{phasetwo}{HTML}{FFF4E0}      %
\definecolor{phasethree}{HTML}{E8F6E8}    %
\begin{tabular}{@{}l p{3.7cm} p{3.7cm} p{3.7cm}@{}}
\toprule
Mode & Phase 1 & Phase 2 & Phase 3 \\
\midrule
\texttt{-fix} & \multicolumn{3}{>{\columncolor{phasefix}}c}{\parbox{12cm}{Pick up the mug from the cabinet, place it under the coffee machine dispenser, then press the start button to brew coffee.}} \\
\midrule
\texttt{-dynamic}
  & \cellcolor{phaseone}Pick up the mug from the cabinet.
  & \cellcolor{phasetwo}Place the mug under the coffee machine dispenser.
  & \cellcolor{phasethree}Press the start button on the coffee machine. \\
\bottomrule
\end{tabular}
\end{table}

\begin{table}[htbp]
\centering
\caption{Composite-task results on 18 RoboCasa365 composites. Phase-$k$ SR counts episodes that reach the $k$-th subgoal milestone.}
\label{tab:composite-aggregate}
\small
\setlength{\tabcolsep}{4pt}
\resizebox{\linewidth}{!}{%
\begin{tabular}{l cc ccc}
\toprule
 & \multicolumn{2}{c}{\texttt{-fix}} & \multicolumn{3}{c}{\texttt{-dynamic}} \\
\cmidrule(lr){2-3} \cmidrule(lr){4-6}
Metric & Baseline & DeMiAn-VLA & Baseline (GT) & DeMiAn-VLA (GT) & DeMiAn-VLA (instr.) \\
\midrule
Mean phase-1 SR   & 50\% & 52\% & 57\% & \textbf{65\%} & 61\% \\
Mean phase-2 SR   & 28\% & \textbf{32\%} & 26\% & 31\% & 30\% \\
Mean full-task SR & 13\% & 15\% & 19\% & \textbf{22\%} & 18\% \\
\bottomrule
\end{tabular}%
}
\end{table}

Table~\ref{tab:composite-aggregate} shows the results of the composite-task chaining experiment.
First, within the OOD \texttt{-fix} format, DeMiAn-VLA improves over the task-only baseline by +2 full-task points (15\% vs 13\%), suggesting that dense annotations yield modest robustness to instruction formats outside training. Second, with GT atomic prompts under \texttt{-dynamic}, DeMiAn-VLA-GT is the strongest configuration overall (22\% vs 19\% baseline-GT), showing that the dense-annotation policy benefits more from subgoal-decomposed prompts than the baseline does. Third, replacing the GT atomic prompt with the DeMiAn instructor at deployment recovers most but not all of this benefit: full-task SR drops to 18\%, sitting between the two GT configurations and approximately 1 point below baseline-GT. The instructor still leads on Phase-1 and Phase-2 milestones (61\%/30\% vs 57\%/26\%), so the gap concentrates in the final phase. Closing this last-phase gap is left for future work.

\subsection{RQ4: Is dense annotation a compute-efficient lever?}
\label{sec:rq4-scaling}
We ask whether the upfront cost of generating dense annotations is recovered by training-time gains, charging caption-generation FLOPs into the compute budget.
\citet{nvidia2025dreamdojo} show that a video-generative backbone trained on human video alone---without any robot action labels---transfers usefully to downstream robot manipulation.
We refer to training a video-generative backbone with human video as the \textit{mid-training} stage, in contrast to the \textit{post-training} stage where an action head is attached to the backbone and robot action data are available. 
We evaluate the effect of DeMiAn across both mid-training and post-training stages.
We first mid-train DeMiAn WAM on 50K human-egocentric clips from EgoVerse~\citep{egoverse} with no action head attached, optimized purely on a video-prediction objective; the DeMiAn annotations enter as language conditioning on the video model. 

\begin{figure}[htbp]
\centering
\begin{minipage}{0.34\textwidth}
\centering
\includegraphics[width=\linewidth]{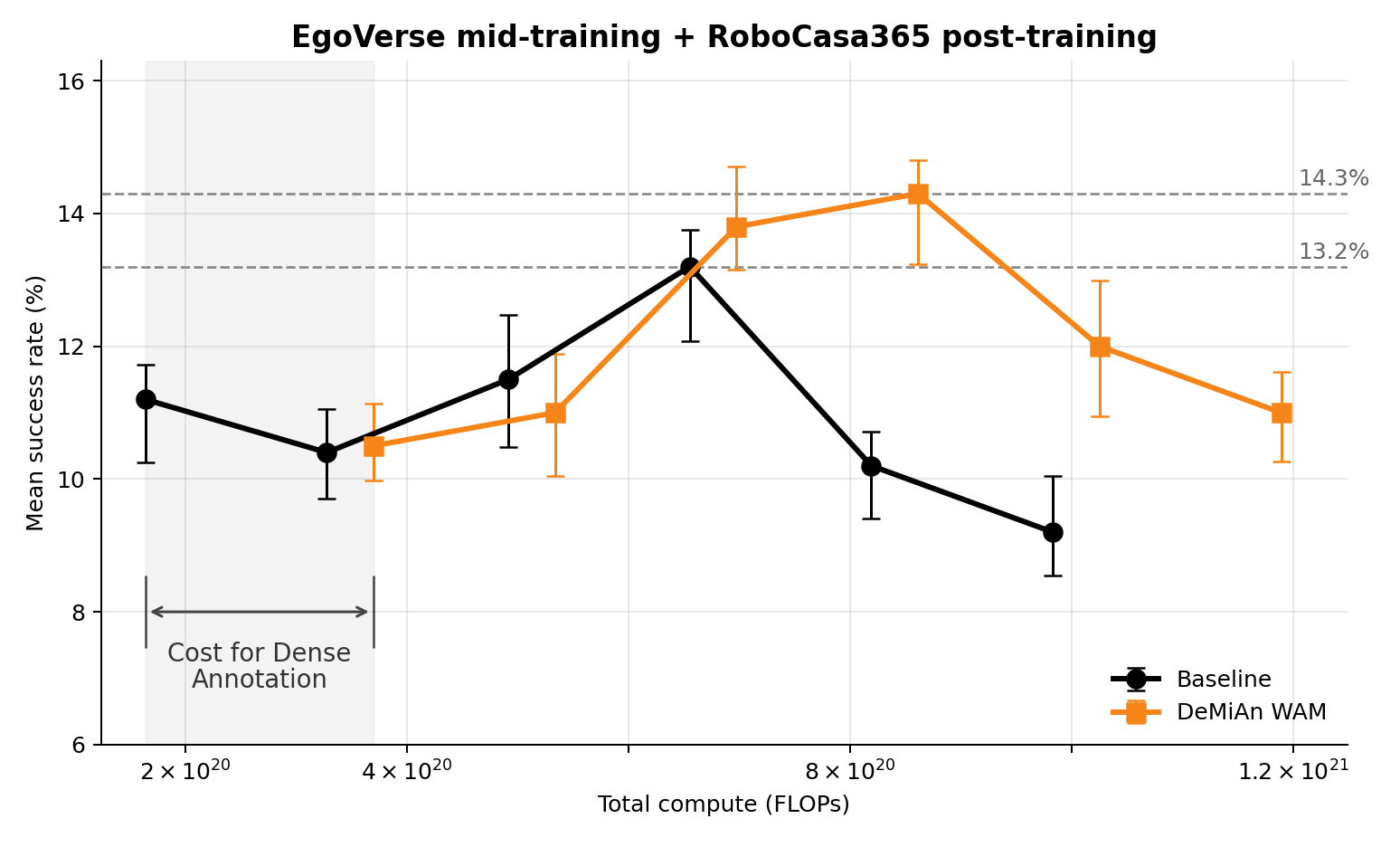}
\end{minipage}
\hfill
\begin{minipage}{0.64\textwidth}
\centering
\includegraphics[width=\linewidth]{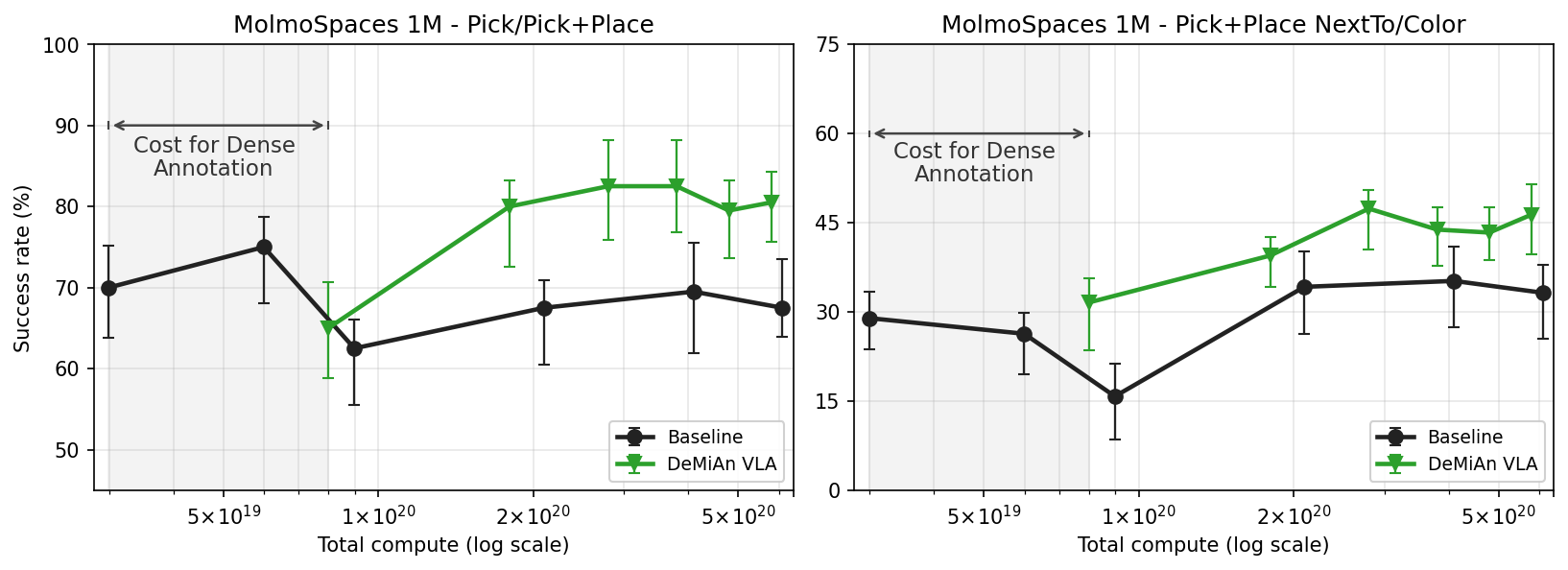}
\end{minipage}
\caption{\textbf{DeMiAn under scaling.} (A)~DeMiAn-WAM mid-training on EgoVerse 50K with dense annotations, evaluated on downstream RoboCasa 365.
The x-axis includes both mid-training compute and annotation-generation compute.
(B)~DeMiAn-VLA post-training on the 1M-scale MolmoBot corpus, evaluated by total success across four MolmoSpaces benchmarks.
The x-axis includes annotation-generation and DeMiAn-VLA post-training FLOPs.}
\label{fig:language-scaling}
\end{figure}

\textbf{Scaling under matched compute.}
We mid-train the Cosmos-Predict 2.5 model on 50K EgoVerse clips with and without dense annotations and post-train DeMiAn-WAM on RoboCasa365 under both conditions.
Figure~\ref{fig:language-scaling} reports both regimes with annotation-generation compute charged into the x-axis. After the small upfront cost of generating annotations, the annotated WAM reaches higher downstream RoboCasa SR than the baseline at nearby compute budgets (Fig.~\ref{fig:language-scaling}A).
On MolmoBot post-training with 1M trajectories, annotation-conditioned policies reach stronger MolmoSpaces SR earlier in training and obtain higher peak performance than the task-only baseline (Fig.~\ref{fig:language-scaling}B).
On the MolmoSpaces NextTo and Color tasks, DeMiAn matches the no-annotation baseline with $\sim$62\% less compute, saving $\sim 1.3\times10^{20}$ FLOPs.

\textbf{Annotation cost.}
A single Qwen3-VL-30B call processes $\approx 8.2$K input tokens (frames + prompt) and emits $\approx 150$ output tokens, costing $\approx \$1.1\times 10^{-3}$ on hosted inference. For a 1M-clip MolmoBot corpus with one DeMiAn aspect per clip, this totals 5.0e19 FLOPs; see Appendix~\ref{sec:annotation-cost-details} for the calculation.

\textbf{Implication.}
At fixed compute, dense annotation accelerates both WAM mid-training and VLA post-training even after charging caption-generation compute against the budget. Combined with the cost asymmetry above, this confirms that adding language to existing demonstrations is a practical and compute-efficient lever in robot policy learning.

%% file: sections/conclusion.tex
\section{Conclusion}
\label{sec:conclusion}

We studied dense, structured language as a supervision channel for robot policy learning, treating annotation \emph{type} as a first-class design choice rather than a fixed prompt format. Our automatic VLM-based pipeline produces four complementary annotation channels---physical motion, scene composition, arm pose, and reasoning---across robot teleoperation and human-egocentric video, spanning RoboCasa 365, MolmoBot, and EgoVerse.

Three findings emerge from this study. First, dense language improves robot manipulation, but the \emph{best} channel is task-dependent: action-, scene-, pose-, and reasoning-style annotations help different task families differently, and a per-task oracle reliably exceeds any single fixed choice. Second, the benefit persists as training scales: dense annotations improve the compute-performance trade-off in both WAM mid-training on EgoVerse and VLA post-training on the 1M-scale MolmoBot corpus, even after charging annotation-generation compute against the budget. Third, dense language is deployable: a small asynchronous instructor closes most of the oracle gap at inference time, and annotation-trained policies generalize to long-horizon composite tasks when paired with subgoal-driven prompt switching.

%% file: sections/appendix.tex
\paragraph{Limitations and future work.}
\label{sec:limitations}
Our annotations are produced by a single frozen VLM and inherit its biases.
Three open problems remain. First, automatically selecting the right
annotation type per task is unsolved: the instructor closes most but not all
of the per-task oracle gap, and on a subset of tasks every fixed aspect
underperforms the task-only baseline. Second, our experiments are simulation
only (RoboCasa~365 and MolmoSpaces); real-robot evaluation would test whether
the language-density lever transfers beyond the visual and dynamics
distributions of these benchmarks. Third, the four-aspect schema is fixed and
chosen heuristically. Natural next steps include scaling beyond a 1M-clip
corpus, mixing annotation channels per segment rather than per dataset, and
tighter coupling between the instructor and the action policy---for example,
training the two jointly so the instructor optimizes directly for downstream
action-policy success rather than imitating the offline reward table.

\section{DeMiAn Multi-Aspect Annotation}
\label{sec:annotation-prompts}

We use Qwen3-VL-30B-A3B-Instruct to generate all four DeMiAn annotation aspects. Each aspect is produced by a separate VLM call conditioned on the segment video, the existing task description, and a context block containing the original task name, available task descriptions, and (for the \texttt{reasoning} aspect on RoboCasa 365) the ground-truth primitive sequence and neighboring segment labels. Each prompt carries a per-aspect length cap. The robot-manipulation datasets (RoboCasa 365 and MolmoSpaces) share one set of prompts written in agent-centric language; EgoVerse uses a separate set adapted to human-egocentric video. We list both sets in the following sections.

\subsection{Robot-manipulation Prompts (RoboCasa 365 and MolmoSpaces)}
\label{sec:annotation-prompts-robot}

RoboCasa 365 and MolmoSpaces share the four agent-centric prompts below. Each is issued as a separate VLM call whose user message bundles the segment video, the task description, and the per-aspect instruction.

\begin{promptbox}{physical\_motion}
Describe the physical movement of the agent. For example, if the agent is moving its arm, describe the movement of the arm. If the agent is moving its hand (or the gripper of a robot arm), describe the movement of the hand or the gripper. If the agent is grasping the object, describe the grasping movement of the gripper or hand. If the agent is moving the object, describe the movement of the object. Focus only on the movements within the given frames. Do not hallucinate or make up the action.
\end{promptbox}

\begin{promptbox}{scene\_composition}
Describe the physical environment shown in the video. List the room type, major fixtures, and visible objects on the surfaces (such as specific food items, appliances, or tools).
\end{promptbox}

\begin{promptbox}{arm\_pose}
Describe the exact physical posture and spatial location of the agent's arm throughout the trajectory. Focus strictly on the arm's pose (posture, gripper state, orientation) relative to the environment at the start, middle, and end of the clip, without describing the action itself.
\end{promptbox}

\begin{promptbox}{reasoning}
Reason about the agent's action and environment in the video clips given the task description. The reasoning should be detailed and specific to the video clips, e.g., why the agent is doing this action, what is the goal of the action, what was the previous action, what was the current action, what should be the next action, is the task completed, etc.
\end{promptbox}

\subsection{Human-Egocentric Video Prompts (EgoVerse)}
\label{sec:annotation-prompts-ego}

For EgoVerse we recast the same four aspects in human-egocentric language and tighten every prompt to a two-sentence cap, since EgoVerse segments are short and the annotations are concatenated with longer narrations elsewhere in the corpus.

\begin{promptbox}{physical\_motion}
Describe the physical movements of the person's hands in this clip. Focus on what each hand is doing: reaching, grasping, lifting, pouring, stirring, placing, etc. Mention the objects being manipulated and the direction of movement. Focus only on the movements within the given frames. Do not hallucinate or make up actions.
\end{promptbox}

\begin{promptbox}{scene\_composition}
Describe the physical environment shown in the images. List the setting, workspace surfaces, and visible objects (such as tools, containers, food items, or appliances). Note their spatial arrangement.
\end{promptbox}

\begin{promptbox}{arm\_pose}
Describe the exact position and posture of the person's hands at the very first frame. What are they holding, touching, or hovering over? Focus strictly on the hands' state relative to the objects and workspace, without describing the action.
\end{promptbox}

\begin{promptbox}{reasoning}
Reason about what the person is doing and why, given the task description and the current action annotation. What is the goal of this action segment? What was likely done before this, and what will likely come next? Is this a preparatory step, the main manipulation, or a cleanup step?
\end{promptbox}

\subsection{Annotation Cost Calculation}
\label{sec:annotation-cost-details}

We use Qwen3-VL-30B-A3B-Instruct through the annotation pipeline.
Each VLM call processes one segment video with a text prompt containing the task description and the requested DeMiAn aspect.
The script defaults to \texttt{max\_tokens=256}; empirically and for the scaling plots we approximate each request as $X_{\mathrm{in}}\approx 8.2\mathrm{K}$ input tokens (video plus prompt) and $X_{\mathrm{out}}\approx 150$ generated tokens. With Qwen3-VL-30B-A3B's MoE routing, we charge only the active parameters, $P_{\mathrm{active}}\approx 3\times 10^9$, giving
\[
F_{\mathrm{call}} \approx 2 P_{\mathrm{active}}(X_{\mathrm{in}} + X_{\mathrm{out}})
\approx 2 \cdot 3\times 10^9 \cdot 8.35\times 10^3
\approx 5.0\times 10^{13}\;\text{FLOPs}.
\]
Thus a one-aspect annotation pass over a 1M-clip MolmoBot corpus costs
\[
F_{\mathrm{cap}} \approx 10^6 F_{\mathrm{call}}
\approx 5.0\times 10^{19}\;\text{FLOPs}.
\]
Using a representative hosted price of \$0.13 per million input tokens and \$0.52 per million output tokens for Qwen3-VL-30B-A3B, this is approximately
\[
10^6\left(8200\cdot \frac{0.13}{10^6} + 150\cdot \frac{0.52}{10^6}\right)
\approx \$1.1\text{K}
\]
for one DeMiAn aspect, or roughly four times larger for annotating all four aspects.

\section{Experiment Details}
\label{sec:experiment-details}

\subsection{Training Details}
\label{sec:training-details}

\textbf{DeMiAn VLA.}
We adopt the \texttt{openpi} model, where the policy applies bidirectional attention over the full prefix of image and language tokens, and is trained with a flow-matching action objective.
When reasoning or dense annotation text is included, it is concatenated to the language prompt and receives no token-level supervision.
We keep the same bidirectional attention pattern, but add an annotation-level language-modeling auxiliary loss with weight $\lambda_{\text{LM}}=0.1$. Let the prefix tokens be
$x = (x^{\text{img}}, x^{\text{task}}, x^{\text{cap}})$, where $x^{\text{cap}}=(c_1,\dots,c_N)$ are the annotation tokens. Let $h_i=f_\theta(x)_i$ be the bidirectional transformer output at annotation position $i$, and let $W$ be the tied LM head. The training objective is
\[
\mathcal{L}
= \mathcal{L}_{\text{FM}} + \lambda_{\text{LM}} \mathcal{L}_{\text{cap}},
\qquad
\mathcal{L}_{\text{cap}}
= -\frac{1}{\sum_{i=1}^{N-1} m_i}
\sum_{i=1}^{N-1} m_i \log p_\theta(c_{i+1}\mid h_i),
\]
where $p_\theta(\cdot\mid h_i)=\mathrm{softmax}(W h_i)$ and $m_i$ is the annotation-token mask.

Because $h_i$ can attend bidirectionally to later annotation tokens $c_{>i}$, $\mathcal{L}_{\text{cap}}$ is not a causal likelihood objective and can be partially solved through attention-based copying.
We therefore use it only as an auxiliary representation-grounding regularizer: it encourages annotation-token hidden states to remain decodable through the LM head, rather than allowing the bidirectional encoder to compress the annotation into a coarse global summary that may discard object identity, spatial relations, or sub-step structure needed by the action expert.
Annotations are produced by an external instructor at inference time, so the model does not generate annotations during deployment and the absence of a causal mask does not create a train-test mismatch.

We post-train two DeMiAn VLA variants from the same backbone, both for 30K steps at global batch size~256 with \texttt{openpi}'s default optimizer and learning-rate schedule.
The RoboCasa variant fine-tunes \texttt{pi05\_base} on the RoboCasa~365 atomic-task split with $\lambda_{\text{LM}}=0.1$.
The MolmoBot variant fine-tunes \texttt{pi05\_droid} on the 20\% subset MolmoBot corpus without $\mathcal{L}_{\text{cap}}$ ($\lambda_{\text{LM}}=0$).
For MolmoBot training, we follow the default training mixture that pairs one primary task family (Pick+Place) with three auxiliary families sampled at relative weights $0.25/0.20/0.15$ for Pick, NextTo, and Color, respectively.

We use two MolmoBot training scales: the main annotation ablations in Tables~\ref{tab:language-study} and~\ref{tab:language-study-molmospace-detail} are run on a 20\% subset (223K trajectories), while the scaling experiment in Section~\ref{sec:rq4-scaling} (Figure~\ref{fig:language-scaling}B) uses the full 1M-clip MolmoBot corpus.

\textbf{DeMiAn WAM.}
\textit{1) Architecture}.
We adopt a prefix-suffix design in the style of GR00T-N1~\citep{nvidia2025gr00tn1openfoundation}, adapted to the Cosmos-Predict 2.5 video DiT backbone, with a joint blockwise attention variant inspired by $\pi_{0.5}$~\cite{black2026pi0visionlanguageactionflowmodel}.
Cosmos Reason 1 \citep{nvidia2025cosmosreason1physicalcommonsense} vision-language encoder jointly tokenizes the current observation frame together with the task description into a unified prefix, which the action head consumes via cross-attention.
The action head is a four-layer cross-attention stack that runs in parallel to the DiT: each of its $K{=}4$ layers taps a DiT block at an evenly spread depth.
Within each layer we apply a single joint attention whose queries come only from the action tokens and whose keys and values are the concatenation of (i) the DiT video features at the tapped depth, (ii) the Reason1 vision-language prefix, and (iii) the action queries themselves.
The joint attention is followed by a position-wise FFN with hidden ratio 4, and both sub-layers are modulated by AdaLN derived from a sinusoidal time embedding that encodes the flow-matching timestep.
Action queries are content-conditioned by a linear projection of the noisy action chunk, and the final action prediction is produced by a zero-initialized linear map onto the 7-DoF action space.

\textit{2) Training}.
We train the action head with rectified-flow matching following $\pi_0$~\cite{black2026pi0visionlanguageactionflowmodel}.
When action labels are available, this loss is combined with the DiT's standard video flow-matching objective so the backbone continues to adapt to the target domain; Reason 1 is kept frozen and used as an online prefix encoder.
We train with AdamW (learning rate $5\times10^{-5}$, weight decay $10^{-5}$), a 3K-step warmup, global batch size 256, and 160K total steps: 80K video-only steps, 50K end-to-end steps on the full dataset, and 30K target-task alignment steps.

\subsection{Evaluation Protocol Details}
\label{sec:eval-protocol-detail}

Table~\ref{tab:language-study} reports RoboCasa results for DeMiAn-VLA, averaged across 5 seeds $\times$ 20 episodes per seed for each task.
MolmoSpaces test results use a 9-benchmark held-out suite and aggregate per-benchmark success rates into task-family averages.

\begin{table}[ht]
\centering
\caption{Evaluation protocol summary across the development and test splits used in the paper.
In RoboCasa, the seed determines all stochastic elements of the environment (object placement, kitchen layout selection, randomized scene elements).}
\label{tab:eval-protocol}
\small
\setlength{\tabcolsep}{3.5pt}
\begin{tabular}{@{}lcccc@{}}
\toprule
Parameter & \shortstack{RoboCasa\\Dev} & \shortstack{RoboCasa\\Test} & \shortstack{MolmoSpaces\\Dev} & \shortstack{MolmoSpaces\\Test} \\
\midrule
Tasks / benchmarks      & 17 atomic      & 17 atomic      & 6 benchmarks   & 9 benchmarks \\
Episodes / unit         & 20             & 100            & 20 (episodes 0–19) & 180 (episodes 20–199) \\
Seed                    & 123-127        & 42-46          & fixed JSON     & fixed JSON \\
Total trials            & 340            & 1{,}700        & 120            & 1,620 \\
Max steps / episode     & 400            & 400            & 607            & 607 \\
Action chunk horizon    & 8              & 8              & 8              & 8 \\
Env split               & target         & target         & bench-v2 dev subset & bench-v2 test suite \\
Camera resolution       & 256$\times$256 & 256$\times$256 & 256$\times$256 & 256$\times$256 \\
\bottomrule
\end{tabular}
\end{table}

\begin{table}[ht]
\centering
\caption{Per-benchmark success rate as a function of training-time annotation.
Values are fractions in $[0, 1]$.
Columns are grouped by task family with calculated \emph{Avg.} columns: Pick averages Std/Hard/OOD, Pick+Place averages Std/Hard, and NextTo averages ID/OOD.
\emph{w/ Instructor} denotes the learned instruction-generation condition.
\emph{Per-task Best (oracle)} takes the column-wise maximum across the annotation rows and serves as an oracle upper bound.}
\label{tab:language-study-molmospace-detail}
\small
\setlength{\tabcolsep}{3.5pt}
\begin{tabular}{l cccc cccc ccc c}
\toprule
 & \multicolumn{4}{c}{Pick} & \multicolumn{4}{c}{Pick+Place} & \multicolumn{3}{c}{NextTo} & Color \\
\cmidrule(lr){2-5} \cmidrule(lr){6-9} \cmidrule(lr){10-12} \cmidrule(lr){13-13}
Annotation type & Std & Hard & OOD & Avg. & Std & Hard & OOD & Avg. & ID & OOD & Avg. & \\
\midrule
\rowcolor{black!8}
\multicolumn{13}{@{}l}{\textbf{DeMiAn VLA}} \\
Baseline                 & .61 & .34 & .18 & .38 & \textbf{.82} & \textbf{.45} & .22 & \textbf{.64} & .25 & .07 & .16 & .41 \\
Physical Motion          & .73 & \textbf{.45} & .24 & .47 & .76 & .44 & .18 & .60 & .20 & \textbf{.13} & .17 & .46 \\
Scene Composition        & .77 & \textbf{.45} & .26 & \textbf{.49} & .80 & .40 & .24 & .60 & .26 & .11 & .19 & .40 \\
Arm Pose                 & .70 & .37 & .18 & .42 & .76 & .44 & .26 & .60 & .18 & .09 & .14 & .45 \\
Reasoning                & .79 & .38 & .26 & .48 & .78 & .41 & .24 & .60 & \textbf{.33} & .12 & \textbf{.23} & .39 \\
\midrule
w/ Instructor            & \textbf{.80} & .40 & \textbf{.28} & \textbf{.49} & .78 & .42 & \textbf{.27} & .60 & .27 & \textbf{.13} & .20 & \textbf{.48} \\
\midrule
Per-task Best (oracle)   & .80 & .45 & .28 & .51 & .82 & .45 & .27 & .64 & .33 & .13 & .23 & .48 \\
\midrule
\rowcolor{black!8}
\multicolumn{13}{@{}l}{\textbf{DeMiAn WAM}} \\
Baseline                 & .40 & .11 & .06 & .19 & .25 & .13 & .02 & .19 & .06 & \textbf{.08} & .07 & .14 \\
Physical Motion          & .44 & \textbf{.15} & .08 & .22 & .27 & .15 & .06 & .21 & .07 & .05 & .06 & .15 \\
Scene Composition        & .45 & .14 & \textbf{.09} & .23 & .26 & .13 & .08 & .20 & .09 & .04 & .07 & .13 \\
Arm Pose                 & .38 & .12 & .06 & .19 & .22 & .12 & .09 & .17 & .06 & .03 & .05 & .15 \\
Reasoning                & .43 & .13 & \textbf{.09} & .22 & .24 & .14 & .08 & .19 & \textbf{.11} & .04 & \textbf{.08} & .13 \\
\midrule
w/ Instructor            & \textbf{.49} & .13 & \textbf{.09} & \textbf{.24} & \textbf{.30} & \textbf{.16} & \textbf{.10} & \textbf{.23} & .09 & .04 & .07 & \textbf{.16} \\
\midrule
Per-task Best (oracle)   & .45 & .15 & .09 & .23 & .27 & .15 & .09 & .21 & .11 & .08 & .10 & .15 \\
\bottomrule
\end{tabular}
\end{table}

\section{Extended Results and Analysis}
\label{sec:more-results-and-analysis}

\subsection{Per-Task and Per-Family Annotation Analysis}
\label{sec:per-task-analysis}

This section expands on the aggregate RQ1 result of Section~\ref{sec:rq1-heterogeneity} (Table~\ref{tab:language-study}) by walking through which annotation aspect wins on which task and how the pattern differs on MolmoSpaces.

\paragraph{Per-aspect strengths on RoboCasa.}
RoboCasa shows structured heterogeneity rather than a single best annotation type.
\texttt{physical\_motion} tends to help contact-changing manipulation, \texttt{scene\_composition} helps tasks that require choosing among similar receptacles or surfaces, and \texttt{arm\_pose} is useful mainly when the task is dominated by body configuration.
By contrast, \texttt{reasoning} has no unique per-task wins under any fixed aspect, suggesting that subgoal-style annotations alone are often dominated by either motion structure or the learned instructor.
The task-only baseline remains strongest on several saturated or annotation-sensitive tasks, so dense annotations are not uniformly beneficial.
The same pattern carries over to DeMiAn WAM at lower absolute SR, while \emph{w/ Instructor} converts these complementary strengths into a single policy and uniquely wins tasks that no fixed aspect dominates.

\paragraph{Per-family patterns on MolmoSpaces.}
For DeMiAn VLA, dense annotations clearly help on \emph{Pick}---every fixed aspect lifts the family average over the baseline, with Scene Composition and the learned instructor tied at the top.
\emph{Pick+Place} is the one family where the task-only baseline holds its ground on average; individual dense aspects carry pockets of strength on the OOD split, but none beat the baseline overall. 
\emph{NextTo} flips the winner to Reasoning, whose subgoal-style framing fits the spatial-relation language this family demands; \emph{Color}, in turn, is uniquely won by the instructor.
The DeMiAn WAM picture is cleaner: the instructor leads on three of the four families (Pick, Pick+Place, Color), and only \emph{NextTo} again favors Reasoning.

Across both backbones, the takeaway is the same---no fixed aspect is universally best, and the per-family heterogeneity is large enough that selecting the aspect at deployment carries real signal.

\subsection{Instructor Training Details}
\label{sec:routing-results}

\begin{table}[ht]
\centering
\caption{\textbf{Results with the Instructor on the RoboCasa Dev set.} 
All variants share the same action policy checkpoint; only the instruction at inference differs. The \emph{Fixed-aspect GT injection} rows (averaged across the 17 RoboCasa atomic tasks) double as an aggregate view of the reward table $w(\tau, k)$ used to train the instructor.}
\label{tab:async-instructor}
\small
\begin{tabular}{lcc}
\toprule
Instruction at inference & SR & $\Delta$ vs baseline \\
\midrule
Baseline (no annotation)                                    & 44.3\%               & ---             \\
\midrule
\multicolumn{3}{@{}l}{\textit{Fixed-aspect GT injection (inference-time analogue of Table~\ref{tab:language-study})}} \\
GT \texttt{physical\_motion}    & 46.1\%               & +1.8            \\
GT \texttt{scene\_composition}  & 48.4\%               & +4.1            \\
GT \texttt{arm\_pose}  & 47.4\%               & +3.1            \\
GT \texttt{reasoning}  & \underline{50.1\%}   & \underline{+5.8} \\
Random per-episode aspect & 46.6\%               & +2.3            \\
\midrule
\multicolumn{3}{@{}l}{\textit{DeMiAn Instructor variants (async, ours)}} \\
\textbf{SFT Instructor}                                      & \textbf{50.4\%}      & \textbf{+6.1}   \\
\midrule
Oracle (per-task-best annotation)                           & 52.4\%               & +8.1            \\
\bottomrule
\end{tabular}
\end{table}

\textbf{Reward table.} The reward table $w(\tau, k)$ records the action policy's dev-set SR on each (task $\tau$, aspect $k$) pair.
We construct it by running the action policy with each of the four GT fixed-aspect annotations across all 17 RoboCasa atomic tasks and recording the resulting SRs.
The aspect-wise aggregate of this table is visible in the \emph{Fixed-aspect GT injection} rows of Table~\ref{tab:async-instructor} (averaged across the 17 tasks); the actual training-time table is per-task and resolves the heterogeneity that this aggregate hides.
Tasks where every aspect underperforms the no-annotation baseline are assigned an empty target, teaching the instructor to abstain.

\textbf{Training procedure.} We fine-tune Qwen3.5-2B with supervised fine-tuning (SFT).
For each training example, we sample a target aspect per task from a softmax over $w(\tau, \cdot)$ (temperature $T{=}2$, top-3 truncation), and use the pipeline-generated annotation for that (task, aspect) pair as the SFT target---this is the \emph{reward-weighted target sampling} setup described in Section~\ref{sec:method}.
The training set contains 3.2K examples.
Training uses 4$\times$GB200 GPUs with effective batch size 256, AdamW with cosine schedule.

\textbf{Ablations.} We also tested Top-1 SFT (target $=$ the single highest-reward aspect per task) and two DPO variants seeded from SFT (cross-aspect preference pairs ordered by $w(\tau, k)$); none outperformed the default at equivalent training budgets, and length-confounded preference pairs caused DPO to drift toward over-long outputs that hurt the action policy. We therefore use SFT with reward-weighted target sampling as the default recipe.

%% file: reference.bib
@inproceedings{walke2023bridgedata,
    title={BridgeData V2: A Dataset for Robot Learning at Scale},
    author={Walke, Homer and Black, Kevin and Lee, Abraham and Kim, Moo Jin and Du, Max and Zheng, Chongyi and Zhao, Tony and Hansen-Estruch, Philippe and Vuong, Quan and He, Andre and Myers, Vivek and Fang, Kuan and Finn, Chelsea and Levine, Sergey},
    booktitle={Conference on Robot Learning (CoRL)},
    year={2023}
}

@misc{open_x_embodiment_rt_x_2023,
title={Open {X-E}mbodiment: Robotic Learning Datasets and {RT-X} Models},
author = {Open X-Embodiment Collaboration and Abby O'Neill and Abdul Rehman and Abhinav Gupta and Abhiram Maddukuri and Abhishek Gupta and Abhishek Padalkar and Abraham Lee and Acorn Pooley and Agrim Gupta and Ajay Mandlekar and Ajinkya Jain and Albert Tung and Alex Bewley and Alex Herzog and Alex Irpan and Alexander Khazatsky and Anant Rai and Anchit Gupta and Andrew Wang and Andrey Kolobov and Anikait Singh and Animesh Garg and Aniruddha Kembhavi and Annie Xie and Anthony Brohan and Antonin Raffin and Archit Sharma and Arefeh Yavary and Arhan Jain and Ashwin Balakrishna and Ayzaan Wahid and Ben Burgess-Limerick and Beomjoon Kim and Bernhard Schölkopf and Blake Wulfe and Brian Ichter and Cewu Lu and Charles Xu and Charlotte Le and Chelsea Finn and Chen Wang and Chenfeng Xu and Cheng Chi and Chenguang Huang and Christine Chan and Christopher Agia and Chuer Pan and Chuyuan Fu and Coline Devin and Danfei Xu and Daniel Morton and Danny Driess and Daphne Chen and Deepak Pathak and Dhruv Shah and Dieter Büchler and Dinesh Jayaraman and Dmitry Kalashnikov and Dorsa Sadigh and Edward Johns and Ethan Foster and Fangchen Liu and Federico Ceola and Fei Xia and Feiyu Zhao and Felipe Vieira Frujeri and Freek Stulp and Gaoyue Zhou and Gaurav S. Sukhatme and Gautam Salhotra and Ge Yan and Gilbert Feng and Giulio Schiavi and Glen Berseth and Gregory Kahn and Guangwen Yang and Guanzhi Wang and Hao Su and Hao-Shu Fang and Haochen Shi and Henghui Bao and Heni Ben Amor and Henrik I Christensen and Hiroki Furuta and Homanga Bharadhwaj and Homer Walke and Hongjie Fang and Huy Ha and Igor Mordatch and Ilija Radosavovic and Isabel Leal and Jacky Liang and Jad Abou-Chakra and Jaehyung Kim and Jaimyn Drake and Jan Peters and Jan Schneider and Jasmine Hsu and Jay Vakil and Jeannette Bohg and Jeffrey Bingham and Jeffrey Wu and Jensen Gao and Jiaheng Hu and Jiajun Wu and Jialin Wu and Jiankai Sun and Jianlan Luo and Jiayuan Gu and Jie Tan and Jihoon Oh and Jimmy Wu and Jingpei Lu and Jingyun Yang and Jitendra Malik and João Silvério and Joey Hejna and Jonathan Booher and Jonathan Tompson and Jonathan Yang and Jordi Salvador and Joseph J. Lim and Junhyek Han and Kaiyuan Wang and Kanishka Rao and Karl Pertsch and Karol Hausman and Keegan Go and Keerthana Gopalakrishnan and Ken Goldberg and Kendra Byrne and Kenneth Oslund and Kento Kawaharazuka and Kevin Black and Kevin Lin and Kevin Zhang and Kiana Ehsani and Kiran Lekkala and Kirsty Ellis and Krishan Rana and Krishnan Srinivasan and Kuan Fang and Kunal Pratap Singh and Kuo-Hao Zeng and Kyle Hatch and Kyle Hsu and Laurent Itti and Lawrence Yunliang Chen and Lerrel Pinto and Li Fei-Fei and Liam Tan and Linxi "Jim" Fan and Lionel Ott and Lisa Lee and Luca Weihs and Magnum Chen and Marion Lepert and Marius Memmel and Masayoshi Tomizuka and Masha Itkina and Mateo Guaman Castro and Max Spero and Maximilian Du and Michael Ahn and Michael C. Yip and Mingtong Zhang and Mingyu Ding and Minho Heo and Mohan Kumar Srirama and Mohit Sharma and Moo Jin Kim and Muhammad Zubair Irshad and Naoaki Kanazawa and Nicklas Hansen and Nicolas Heess and Nikhil J Joshi and Niko Suenderhauf and Ning Liu and Norman Di Palo and Nur Muhammad Mahi Shafiullah and Oier Mees and Oliver Kroemer and Osbert Bastani and Pannag R Sanketi and Patrick "Tree" Miller and Patrick Yin and Paul Wohlhart and Peng Xu and Peter David Fagan and Peter Mitrano and Pierre Sermanet and Pieter Abbeel and Priya Sundaresan and Qiuyu Chen and Quan Vuong and Rafael Rafailov and Ran Tian and Ria Doshi and Roberto Mart{'i}n-Mart{'i}n and Rohan Baijal and Rosario Scalise and Rose Hendrix and Roy Lin and Runjia Qian and Ruohan Zhang and Russell Mendonca and Rutav Shah and Ryan Hoque and Ryan Julian and Samuel Bustamante and Sean Kirmani and Sergey Levine and Shan Lin and Sherry Moore and Shikhar Bahl and Shivin Dass and Shubham Sonawani and Shubham Tulsiani and Shuran Song and Sichun Xu and Siddhant Haldar and Siddharth Karamcheti and Simeon Adebola and Simon Guist and Soroush Nasiriany and Stefan Schaal and Stefan Welker and Stephen Tian and Subramanian Ramamoorthy and Sudeep Dasari and Suneel Belkhale and Sungjae Park and Suraj Nair and Suvir Mirchandani and Takayuki Osa and Tanmay Gupta and Tatsuya Harada and Tatsuya Matsushima and Ted Xiao and Thomas Kollar and Tianhe Yu and Tianli Ding and Todor Davchev and Tony Z. Zhao and Travis Armstrong and Trevor Darrell and Trinity Chung and Vidhi Jain and Vikash Kumar and Vincent Vanhoucke and Vitor Guizilini and Wei Zhan and Wenxuan Zhou and Wolfram Burgard and Xi Chen and Xiangyu Chen and Xiaolong Wang and Xinghao Zhu and Xinyang Geng and Xiyuan Liu and Xu Liangwei and Xuanlin Li and Yansong Pang and Yao Lu and Yecheng Jason Ma and Yejin Kim and Yevgen Chebotar and Yifan Zhou and Yifeng Zhu and Yilin Wu and Ying Xu and Yixuan Wang and Yonatan Bisk and Yongqiang Dou and Yoonyoung Cho and Youngwoon Lee and Yuchen Cui and Yue Cao and Yueh-Hua Wu and Yujin Tang and Yuke Zhu and Yunchu Zhang and Yunfan Jiang and Yunshuang Li and Yunzhu Li and Yusuke Iwasawa and Yutaka Matsuo and Zehan Ma and Zhuo Xu and Zichen Jeff Cui and Zichen Zhang and Zipeng Fu and Zipeng Lin},
howpublished  = {\url{https://arxiv.org/abs/2310.08864}},
year = {2023},
}

@misc{khazatsky2025droidlargescaleinthewildrobot,
      title={DROID: A Large-Scale In-The-Wild Robot Manipulation Dataset}, 
      author={Alexander Khazatsky and Karl Pertsch and Suraj Nair and Ashwin Balakrishna and Sudeep Dasari and Siddharth Karamcheti and Soroush Nasiriany and Mohan Kumar Srirama and Lawrence Yunliang Chen and Kirsty Ellis and Peter David Fagan and Joey Hejna and Masha Itkina and Marion Lepert and Yecheng Jason Ma and Patrick Tree Miller and Jimmy Wu and Suneel Belkhale and Shivin Dass and Huy Ha and Arhan Jain and Abraham Lee and Youngwoon Lee and Marius Memmel and Sungjae Park and Ilija Radosavovic and Kaiyuan Wang and Albert Zhan and Kevin Black and Cheng Chi and Kyle Beltran Hatch and Shan Lin and Jingpei Lu and Jean Mercat and Abdul Rehman and Pannag R Sanketi and Archit Sharma and Cody Simpson and Quan Vuong and Homer Rich Walke and Blake Wulfe and Ted Xiao and Jonathan Heewon Yang and Arefeh Yavary and Tony Z. Zhao and Christopher Agia and Rohan Baijal and Mateo Guaman Castro and Daphne Chen and Qiuyu Chen and Trinity Chung and Jaimyn Drake and Ethan Paul Foster and Jensen Gao and Vitor Guizilini and David Antonio Herrera and Minho Heo and Kyle Hsu and Jiaheng Hu and Muhammad Zubair Irshad and Donovon Jackson and Charlotte Le and Yunshuang Li and Kevin Lin and Roy Lin and Zehan Ma and Abhiram Maddukuri and Suvir Mirchandani and Daniel Morton and Tony Nguyen and Abigail O'Neill and Rosario Scalise and Derick Seale and Victor Son and Stephen Tian and Emi Tran and Andrew E. Wang and Yilin Wu and Annie Xie and Jingyun Yang and Patrick Yin and Yunchu Zhang and Osbert Bastani and Glen Berseth and Jeannette Bohg and Ken Goldberg and Abhinav Gupta and Abhishek Gupta and Dinesh Jayaraman and Joseph J Lim and Jitendra Malik and Roberto Martín-Martín and Subramanian Ramamoorthy and Dorsa Sadigh and Shuran Song and Jiajun Wu and Michael C. Yip and Yuke Zhu and Thomas Kollar and Sergey Levine and Chelsea Finn},
      year={2025},
      eprint={2403.12945},
      archivePrefix={arXiv},
      primaryClass={cs.RO},
      url={https://arxiv.org/abs/2403.12945}, 
}

@misc{egoverse,
      title={EgoVerse: An Egocentric Human Dataset for Robot Learning from Around the World}, 
      author={Ryan Punamiya and Simar Kareer and Zeyi Liu and Josh Citron and Ri-Zhao Qiu and Xiongyi Cai and Alexey Gavryushin and Jiaqi Chen and Davide Liconti and Lawrence Y. Zhu and Patcharapong Aphiwetsa and Baoyu Li and Aniketh Cheluva and Pranav Kuppili and Yangcen Liu and Dhruv Patel and Aidan Gao and Hye-Young Chung and Ryan Co and Renee Zbizika and Jeff Liu and Xiaomeng Xu and Haoyu Xiong and Geng Chen and Sebastiano Oliani and Chenyu Yang and Xi Wang and James Fort and Richard Newcombe and Josh Gao and Jason Chong and Garrett Matsuda and Aseem Doriwala and Marc Pollefeys and Robert Katzschmann and Xiaolong Wang and Shuran Song and Judy Hoffman and Danfei Xu},
      year={2026},
      eprint={2604.07607},
      archivePrefix={arXiv},
      primaryClass={cs.RO},
      url={https://arxiv.org/abs/2604.07607}, 
}

@misc{robocasa365,
      title={RoboCasa365: A Large-Scale Simulation Framework for Training and Benchmarking Generalist Robots}, 
      author={Soroush Nasiriany and Sepehr Nasiriany and Abhiram Maddukuri and Yuke Zhu},
      year={2026},
      eprint={2603.04356},
      archivePrefix={arXiv},
      primaryClass={cs.RO},
      url={https://arxiv.org/abs/2603.04356}, 
}

@misc{robocasa,
      title={RoboCasa: Large-Scale Simulation of Everyday Tasks for Generalist Robots}, 
      author={Soroush Nasiriany and Abhiram Maddukuri and Lance Zhang and Adeet Parikh and Aaron Lo and Abhishek Joshi and Ajay Mandlekar and Yuke Zhu},
      year={2024},
      eprint={2406.02523},
      archivePrefix={arXiv},
      primaryClass={cs.RO},
      url={https://arxiv.org/abs/2406.02523}, 
}

@article{Brohan2022RT1RT,
  title={RT-1: Robotics Transformer for Real-World Control at Scale},
  author={Anthony Brohan and Noah Brown and Justice Carbajal and Yevgen Chebotar and Joseph Dabis and Chelsea Finn and Keerthana Gopalakrishnan and Karol Hausman and Alexander Herzog and Jasmine Hsu and Julian Ibarz and Brian Ichter and Alex Irpan and Tomas Jackson and Sally Jesmonth and Nikhil J. Joshi and Ryan C. Julian and Dmitry Kalashnikov and Yuheng Kuang and Isabel Leal and Kuang-Huei Lee and Sergey Levine and Yao Lu and Utsav Malla and Deeksha Manjunath and Igor Mordatch and Ofir Nachum and Carolina Parada and Jodilyn Peralta and Emily Perez and Karl Pertsch and Jornell Quiambao and Kanishka Rao and Michael S. Ryoo and Grecia Salazar and Pannag R. Sanketi and Kevin Sayed and Jaspiar Singh and Sumedh Anand Sontakke and Austin Stone and Clayton Tan and Huong Tran and Vincent Vanhoucke and Steve Vega and Quan Ho Vuong and F. Xia and Ted Xiao and Peng Xu and Sichun Xu and Tianhe Yu and Brianna Zitkovich},
  journal={ArXiv},
  year={2022},
  volume={abs/2212.06817},
  url={https://api.semanticscholar.org/CorpusID:254591260}
}

@InProceedings{pmlr-v229-zitkovich23a,
  title = 	 {RT-2: Vision-Language-Action Models Transfer Web Knowledge to Robotic Control},
  author =       {Zitkovich, Brianna and Yu, Tianhe and Xu, Sichun and Xu, Peng and Xiao, Ted and Xia, Fei and Wu, Jialin and Wohlhart, Paul and Welker, Stefan and Wahid, Ayzaan and Vuong, Quan and Vanhoucke, Vincent and Tran, Huong and Soricut, Radu and Singh, Anikait and Singh, Jaspiar and Sermanet, Pierre and Sanketi, Pannag R. and Salazar, Grecia and Ryoo, Michael S. and Reymann, Krista and Rao, Kanishka and Pertsch, Karl and Mordatch, Igor and Michalewski, Henryk and Lu, Yao and Levine, Sergey and Lee, Lisa and Lee, Tsang-Wei Edward and Leal, Isabel and Kuang, Yuheng and Kalashnikov, Dmitry and Julian, Ryan and Joshi, Nikhil J. and Irpan, Alex and Ichter, Brian and Hsu, Jasmine and Herzog, Alexander and Hausman, Karol and Gopalakrishnan, Keerthana and Fu, Chuyuan and Florence, Pete and Finn, Chelsea and Dubey, Kumar Avinava and Driess, Danny and Ding, Tianli and Choromanski, Krzysztof Marcin and Chen, Xi and Chebotar, Yevgen and Carbajal, Justice and Brown, Noah and Brohan, Anthony and Arenas, Montserrat Gonzalez and Han, Kehang},
  booktitle = 	 {Proceedings of The 7th Conference on Robot Learning},
  pages = 	 {2165--2183},
  year = 	 {2023},
  editor = 	 {Tan, Jie and Toussaint, Marc and Darvish, Kourosh},
  volume = 	 {229},
  series = 	 {Proceedings of Machine Learning Research},
  month = 	 {06--09 Nov},
  publisher =    {PMLR},
  url = 	 {https://proceedings.mlr.press/v229/zitkovich23a.html}
}

@misc{black2026pi0visionlanguageactionflowmodel,
      title={$\pi_0$: A Vision-Language-Action Flow Model for General Robot Control}, 
      author={Kevin Black and Noah Brown and Danny Driess and Adnan Esmail and Michael Equi and Chelsea Finn and Niccolo Fusai and Lachy Groom and Karol Hausman and Brian Ichter and Szymon Jakubczak and Tim Jones and Liyiming Ke and Sergey Levine and Adrian Li-Bell and Mohith Mothukuri and Suraj Nair and Karl Pertsch and Lucy Xiaoyang Shi and James Tanner and Quan Vuong and Anna Walling and Haohuan Wang and Ury Zhilinsky},
      year={2026},
      eprint={2410.24164},
      archivePrefix={arXiv},
      primaryClass={cs.LG},
      url={https://arxiv.org/abs/2410.24164}, 
}

@InProceedings{pmlr-v305-black25a,
  title = 	 {$\pi_{0.5}$: a Vision-Language-Action Model with Open-World Generalization},
  author =       {Black, Kevin and Brown, Noah and Darpinian, James and Dhabalia, Karan and Driess, Danny and Esmail, Adnan and Equi, Michael Robert and Finn, Chelsea and Fusai, Niccolo and Galliker, Manuel Y. and Ghosh, Dibya and Groom, Lachy and Hausman, Karol and ichter, brian and Jakubczak, Szymon and Jones, Tim and Ke, Liyiming and LeBlanc, Devin and Levine, Sergey and Li-Bell, Adrian and Mothukuri, Mohith and Nair, Suraj and Pertsch, Karl and Ren, Allen Z. and Shi, Lucy Xiaoyang and Smith, Laura and Springenberg, Jost Tobias and Stachowicz, Kyle and Tanner, James and Vuong, Quan and Walke, Homer and Walling, Anna and Wang, Haohuan and Yu, Lili and Zhilinsky, Ury},
  booktitle = 	 {Proceedings of The 9th Conference on Robot Learning},
  pages = 	 {17--40},
  year = 	 {2025},
  editor = 	 {Lim, Joseph and Song, Shuran and Park, Hae-Won},
  volume = 	 {305},
  series = 	 {Proceedings of Machine Learning Research},
  month = 	 {27--30 Sep},
  publisher =    {PMLR},
  url = 	 {https://proceedings.mlr.press/v305/black25a.html}
}

@misc{bai2025qwen3vltechnicalreport,
      title={Qwen3-VL Technical Report}, 
      author={Shuai Bai and Yuxuan Cai and Ruizhe Chen and Keqin Chen and Xionghui Chen and Zesen Cheng and Lianghao Deng and Wei Ding and Chang Gao and Chunjiang Ge and Wenbin Ge and Zhifang Guo and Qidong Huang and Jie Huang and Fei Huang and Binyuan Hui and Shutong Jiang and Zhaohai Li and Mingsheng Li and Mei Li and Kaixin Li and Zicheng Lin and Junyang Lin and Xuejing Liu and Jiawei Liu and Chenglong Liu and Yang Liu and Dayiheng Liu and Shixuan Liu and Dunjie Lu and Ruilin Luo and Chenxu Lv and Rui Men and Lingchen Meng and Xuancheng Ren and Xingzhang Ren and Sibo Song and Yuchong Sun and Jun Tang and Jianhong Tu and Jianqiang Wan and Peng Wang and Pengfei Wang and Qiuyue Wang and Yuxuan Wang and Tianbao Xie and Yiheng Xu and Haiyang Xu and Jin Xu and Zhibo Yang and Mingkun Yang and Jianxin Yang and An Yang and Bowen Yu and Fei Zhang and Hang Zhang and Xi Zhang and Bo Zheng and Humen Zhong and Jingren Zhou and Fan Zhou and Jing Zhou and Yuanzhi Zhu and Ke Zhu},
      year={2025},
      eprint={2511.21631},
      archivePrefix={arXiv},
      primaryClass={cs.CV},
      url={https://arxiv.org/abs/2511.21631}, 
}

@misc{shridhar2022cliport,
      title={CLIPort: What and Where Pathways for Robotic Manipulation}, 
      author={Mohit Shridhar and Lucas Manuelli and Dieter Fox},
      year={2021},
      eprint={2109.12098},
      archivePrefix={arXiv},
      primaryClass={cs.RO},
      url={https://arxiv.org/abs/2109.12098}, 
}

@misc{radford2021clip,
      title={Learning Transferable Visual Models From Natural Language Supervision}, 
      author={Alec Radford and Jong Wook Kim and Chris Hallacy and Aditya Ramesh and Gabriel Goh and Sandhini Agarwal and Girish Sastry and Amanda Askell and Pamela Mishkin and Jack Clark and Gretchen Krueger and Ilya Sutskever},
      year={2021},
      eprint={2103.00020},
      archivePrefix={arXiv},
      primaryClass={cs.CV},
      url={https://arxiv.org/abs/2103.00020}, 
}

@misc{jang2022bcz,
      title={BC-Z: Zero-Shot Task Generalization with Robotic Imitation Learning}, 
      author={Eric Jang and Alex Irpan and Mohi Khansari and Daniel Kappler and Frederik Ebert and Corey Lynch and Sergey Levine and Chelsea Finn},
      year={2022},
      eprint={2202.02005},
      archivePrefix={arXiv},
      primaryClass={cs.RO},
      url={https://arxiv.org/abs/2202.02005}, 
}

@misc{jiang2023vima,
      title={VIMA: General Robot Manipulation with Multimodal Prompts}, 
      author={Yunfan Jiang and Agrim Gupta and Zichen Zhang and Guanzhi Wang and Yongqiang Dou and Yanjun Chen and Li Fei-Fei and Anima Anandkumar and Yuke Zhu and Linxi Fan},
      year={2023},
      eprint={2210.03094},
      archivePrefix={arXiv},
      primaryClass={cs.RO},
      url={https://arxiv.org/abs/2210.03094}, 
}

@misc{ahn2022icanisay,
      title={Do As I Can, Not As I Say: Grounding Language in Robotic Affordances}, 
      author={Michael Ahn and Anthony Brohan and Noah Brown and Yevgen Chebotar and Omar Cortes and Byron David and Chelsea Finn and Chuyuan Fu and Keerthana Gopalakrishnan and Karol Hausman and Alex Herzog and Daniel Ho and Jasmine Hsu and Julian Ibarz and Brian Ichter and Alex Irpan and Eric Jang and Rosario Jauregui Ruano and Kyle Jeffrey and Sally Jesmonth and Nikhil J Joshi and Ryan Julian and Dmitry Kalashnikov and Yuheng Kuang and Kuang-Huei Lee and Sergey Levine and Yao Lu and Linda Luu and Carolina Parada and Peter Pastor and Jornell Quiambao and Kanishka Rao and Jarek Rettinghouse and Diego Reyes and Pierre Sermanet and Nicolas Sievers and Clayton Tan and Alexander Toshev and Vincent Vanhoucke and Fei Xia and Ted Xiao and Peng Xu and Sichun Xu and Mengyuan Yan and Andy Zeng},
      year={2022},
      eprint={2204.01691},
      archivePrefix={arXiv},
      primaryClass={cs.RO},
      url={https://arxiv.org/abs/2204.01691}, 
}

@misc{driess2023palme,
      title={PaLM-E: An Embodied Multimodal Language Model}, 
      author={Danny Driess and Fei Xia and Mehdi S. M. Sajjadi and Corey Lynch and Aakanksha Chowdhery and Brian Ichter and Ayzaan Wahid and Jonathan Tompson and Quan Vuong and Tianhe Yu and Wenlong Huang and Yevgen Chebotar and Pierre Sermanet and Daniel Duckworth and Sergey Levine and Vincent Vanhoucke and Karol Hausman and Marc Toussaint and Klaus Greff and Andy Zeng and Igor Mordatch and Pete Florence},
      year={2023},
      eprint={2303.03378},
      archivePrefix={arXiv},
      primaryClass={cs.LG},
      url={https://arxiv.org/abs/2303.03378}, 
}

@misc{brohan2023rt2,
      title={RT-2: Vision-Language-Action Models Transfer Web Knowledge to Robotic Control}, 
      author={Anthony Brohan and Noah Brown and Justice Carbajal and Yevgen Chebotar and Xi Chen and Krzysztof Choromanski and Tianli Ding and Danny Driess and Avinava Dubey and Chelsea Finn and Pete Florence and Chuyuan Fu and Montse Gonzalez Arenas and Keerthana Gopalakrishnan and Kehang Han and Karol Hausman and Alexander Herzog and Jasmine Hsu and Brian Ichter and Alex Irpan and Nikhil Joshi and Ryan Julian and Dmitry Kalashnikov and Yuheng Kuang and Isabel Leal and Lisa Lee and Tsang-Wei Edward Lee and Sergey Levine and Yao Lu and Henryk Michalewski and Igor Mordatch and Karl Pertsch and Kanishka Rao and Krista Reymann and Michael Ryoo and Grecia Salazar and Pannag Sanketi and Pierre Sermanet and Jaspiar Singh and Anikait Singh and Radu Soricut and Huong Tran and Vincent Vanhoucke and Quan Vuong and Ayzaan Wahid and Stefan Welker and Paul Wohlhart and Jialin Wu and Fei Xia and Ted Xiao and Peng Xu and Sichun Xu and Tianhe Yu and Brianna Zitkovich},
      year={2023},
      eprint={2307.15818},
      archivePrefix={arXiv},
      primaryClass={cs.RO},
      url={https://arxiv.org/abs/2307.15818}, 
}

@misc{kim2024openvla,
      title={OpenVLA: An Open-Source Vision-Language-Action Model}, 
      author={Moo Jin Kim and Karl Pertsch and Siddharth Karamcheti and Ted Xiao and Ashwin Balakrishna and Suraj Nair and Rafael Rafailov and Ethan Foster and Grace Lam and Pannag Sanketi and Quan Vuong and Thomas Kollar and Benjamin Burchfiel and Russ Tedrake and Dorsa Sadigh and Sergey Levine and Percy Liang and Chelsea Finn},
      year={2024},
      eprint={2406.09246},
      archivePrefix={arXiv},
      primaryClass={cs.RO},
      url={https://arxiv.org/abs/2406.09246}, 
}

@misc{belkhale2024rth,
      title={RT-H: Action Hierarchies Using Language}, 
      author={Suneel Belkhale and Tianli Ding and Ted Xiao and Pierre Sermanet and Quon Vuong and Jonathan Tompson and Yevgen Chebotar and Debidatta Dwibedi and Dorsa Sadigh},
      year={2024},
      eprint={2403.01823},
      archivePrefix={arXiv},
      primaryClass={cs.RO},
      url={https://arxiv.org/abs/2403.01823}, 
}

@misc{zawalski2024ecot,
      title={Robotic Control via Embodied Chain-of-Thought Reasoning}, 
      author={Michał Zawalski and William Chen and Karl Pertsch and Oier Mees and Chelsea Finn and Sergey Levine},
      year={2025},
      eprint={2407.08693},
      archivePrefix={arXiv},
      primaryClass={cs.RO},
      url={https://arxiv.org/abs/2407.08693}, 
}

@misc{wei2022cot,
      title={Chain-of-Thought Prompting Elicits Reasoning in Large Language Models}, 
      author={Jason Wei and Xuezhi Wang and Dale Schuurmans and Maarten Bosma and Brian Ichter and Fei Xia and Ed Chi and Quoc Le and Denny Zhou},
      year={2023},
      eprint={2201.11903},
      archivePrefix={arXiv},
      primaryClass={cs.CL},
      url={https://arxiv.org/abs/2201.11903}, 
}

@misc{huang2023inner,
      title={Inner Monologue: Embodied Reasoning through Planning with Language Models}, 
      author={Wenlong Huang and Fei Xia and Ted Xiao and Harris Chan and Jacky Liang and Pete Florence and Andy Zeng and Jonathan Tompson and Igor Mordatch and Yevgen Chebotar and Pierre Sermanet and Noah Brown and Tomas Jackson and Linda Luu and Sergey Levine and Karol Hausman and Brian Ichter},
      year={2022},
      eprint={2207.05608},
      archivePrefix={arXiv},
      primaryClass={cs.RO},
      url={https://arxiv.org/abs/2207.05608}, 
}

@misc{liang2023code,
      title={Code as Policies: Language Model Programs for Embodied Control}, 
      author={Jacky Liang and Wenlong Huang and Fei Xia and Peng Xu and Karol Hausman and Brian Ichter and Pete Florence and Andy Zeng},
      year={2023},
      eprint={2209.07753},
      archivePrefix={arXiv},
      primaryClass={cs.RO},
      url={https://arxiv.org/abs/2209.07753}, 
}

@misc{huang2022language,
      title={Language Models as Zero-Shot Planners: Extracting Actionable Knowledge for Embodied Agents}, 
      author={Wenlong Huang and Pieter Abbeel and Deepak Pathak and Igor Mordatch},
      year={2022},
      eprint={2201.07207},
      archivePrefix={arXiv},
      primaryClass={cs.LG},
      url={https://arxiv.org/abs/2201.07207}, 
}

@misc{singh2023progprompt,
      title={ProgPrompt: Generating Situated Robot Task Plans using Large Language Models}, 
      author={Ishika Singh and Valts Blukis and Arsalan Mousavian and Ankit Goyal and Danfei Xu and Jonathan Tremblay and Dieter Fox and Jesse Thomason and Animesh Garg},
      year={2022},
      eprint={2209.11302},
      archivePrefix={arXiv},
      primaryClass={cs.RO},
      url={https://arxiv.org/abs/2209.11302}, 
}

@misc{nair2022r3m,
      title={R3M: A Universal Visual Representation for Robot Manipulation}, 
      author={Suraj Nair and Aravind Rajeswaran and Vikash Kumar and Chelsea Finn and Abhinav Gupta},
      year={2022},
      eprint={2203.12601},
      archivePrefix={arXiv},
      primaryClass={cs.RO},
      url={https://arxiv.org/abs/2203.12601}, 
}

@misc{karamcheti2023voltron,
      title={Language-Driven Representation Learning for Robotics}, 
      author={Siddharth Karamcheti and Suraj Nair and Annie S. Chen and Thomas Kollar and Chelsea Finn and Dorsa Sadigh and Percy Liang},
      year={2023},
      eprint={2302.12766},
      archivePrefix={arXiv},
      primaryClass={cs.RO},
      url={https://arxiv.org/abs/2302.12766}, 
}

@misc{ma2023liv,
      title={LIV: Language-Image Representations and Rewards for Robotic Control}, 
      author={Yecheng Jason Ma and William Liang and Vaidehi Som and Vikash Kumar and Amy Zhang and Osbert Bastani and Dinesh Jayaraman},
      year={2023},
      eprint={2306.00958},
      archivePrefix={arXiv},
      primaryClass={cs.RO},
      url={https://arxiv.org/abs/2306.00958}, 
}

@misc{yu2023rosie,
      title={Scaling Robot Learning with Semantically Imagined Experience}, 
      author={Tianhe Yu and Ted Xiao and Austin Stone and Jonathan Tompson and Anthony Brohan and Su Wang and Jaspiar Singh and Clayton Tan and Dee M and Jodilyn Peralta and Brian Ichter and Karol Hausman and Fei Xia},
      year={2023},
      eprint={2302.11550},
      archivePrefix={arXiv},
      primaryClass={cs.RO},
      url={https://arxiv.org/abs/2302.11550}, 
}

@misc{hu2023thoughtcloning,
      title={Thought Cloning: Learning to Think while Acting by Imitating Human Thinking}, 
      author={Shengran Hu and Jeff Clune},
      year={2024},
      eprint={2306.00323},
      archivePrefix={arXiv},
      primaryClass={cs.AI},
      url={https://arxiv.org/abs/2306.00323}, 
}

@misc{nvidia2025dreamdojo,
      title={DreamDojo: A Generalist Robot World Model from Large-Scale Human Videos}, 
      author={Shenyuan Gao and William Liang and Kaiyuan Zheng and Ayaan Malik and Seonghyeon Ye and Sihyun Yu and Wei-Cheng Tseng and Yuzhu Dong and Kaichun Mo and Chen-Hsuan Lin and Qianli Ma and Seungjun Nah and Loic Magne and Jiannan Xiang and Yuqi Xie and Ruijie Zheng and Dantong Niu and You Liang Tan and K. R. Zentner and George Kurian and Suneel Indupuru and Pooya Jannaty and Jinwei Gu and Jun Zhang and Jitendra Malik and Pieter Abbeel and Ming-Yu Liu and Yuke Zhu and Joel Jang and Linxi "Jim" Fan},
      year={2026},
      eprint={2602.06949},
      archivePrefix={arXiv},
      primaryClass={cs.RO},
      url={https://arxiv.org/abs/2602.06949}, 
}

@misc{gear2025dreamzero,
      title={World Action Models are Zero-shot Policies}, 
      author={Seonghyeon Ye and Yunhao Ge and Kaiyuan Zheng and Shenyuan Gao and Sihyun Yu and George Kurian and Suneel Indupuru and You Liang Tan and Chuning Zhu and Jiannan Xiang and Ayaan Malik and Kyungmin Lee and William Liang and Nadun Ranawaka and Jiasheng Gu and Yinzhen Xu and Guanzhi Wang and Fengyuan Hu and Avnish Narayan and Johan Bjorck and Jing Wang and Gwanghyun Kim and Dantong Niu and Ruijie Zheng and Yuqi Xie and Jimmy Wu and Qi Wang and Ryan Julian and Danfei Xu and Yilun Du and Yevgen Chebotar and Scott Reed and Jan Kautz and Yuke Zhu and Linxi "Jim" Fan and Joel Jang},
      year={2026},
      eprint={2602.15922},
      archivePrefix={arXiv},
      primaryClass={cs.RO},
      url={https://arxiv.org/abs/2602.15922}, 
}

@misc{nvidia2025cosmos_predict,
      title={World Simulation with Video Foundation Models for Physical AI}, 
      author={NVIDIA and : and Arslan Ali and Junjie Bai and Maciej Bala and Yogesh Balaji and Aaron Blakeman and Tiffany Cai and Jiaxin Cao and Tianshi Cao and Elizabeth Cha and Yu-Wei Chao and Prithvijit Chattopadhyay and Mike Chen and Yongxin Chen and Yu Chen and Shuai Cheng and Yin Cui and Jenna Diamond and Yifan Ding and Jiaojiao Fan and Linxi Fan and Liang Feng and Francesco Ferroni and Sanja Fidler and Xiao Fu and Ruiyuan Gao and Yunhao Ge and Jinwei Gu and Aryaman Gupta and Siddharth Gururani and Imad El Hanafi and Ali Hassani and Zekun Hao and Jacob Huffman and Joel Jang and Pooya Jannaty and Jan Kautz and Grace Lam and Xuan Li and Zhaoshuo Li and Maosheng Liao and Chen-Hsuan Lin and Tsung-Yi Lin and Yen-Chen Lin and Huan Ling and Ming-Yu Liu and Xian Liu and Yifan Lu and Alice Luo and Qianli Ma and Hanzi Mao and Kaichun Mo and Seungjun Nah and Yashraj Narang and Abhijeet Panaskar and Lindsey Pavao and Trung Pham and Morteza Ramezanali and Fitsum Reda and Scott Reed and Xuanchi Ren and Haonan Shao and Yue Shen and Stella Shi and Shuran Song and Bartosz Stefaniak and Shangkun Sun and Shitao Tang and Sameena Tasmeen and Lyne Tchapmi and Wei-Cheng Tseng and Jibin Varghese and Andrew Z. Wang and Hao Wang and Haoxiang Wang and Heng Wang and Ting-Chun Wang and Fangyin Wei and Jiashu Xu and Dinghao Yang and Xiaodong Yang and Haotian Ye and Seonghyeon Ye and Xiaohui Zeng and Jing Zhang and Qinsheng Zhang and Kaiwen Zheng and Andrew Zhu and Yuke Zhu},
      year={2026},
      eprint={2511.00062},
      archivePrefix={arXiv},
      primaryClass={cs.CV},
      url={https://arxiv.org/abs/2511.00062}, 
}

@misc{kim2026molmospaces,
      title={MolmoSpaces: A Large-Scale Open Ecosystem for Robot Navigation and Manipulation}, 
      author={Yejin Kim and Wilbert Pumacay and Omar Rayyan and Max Argus and Winson Han and Eli VanderBilt and Jordi Salvador and Abhay Deshpande and Rose Hendrix and Snehal Jauhri and Shuo Liu and Nur Muhammad Mahi Shafiullah and Maya Guru and Ainaz Eftekhar and Karen Farley and Donovan Clay and Jiafei Duan and Arjun Guru and Piper Wolters and Alvaro Herrasti and Ying-Chun Lee and Georgia Chalvatzaki and Yuchen Cui and Ali Farhadi and Dieter Fox and Ranjay Krishna},
      year={2026},
      eprint={2602.11337},
      archivePrefix={arXiv},
      primaryClass={cs.RO},
      url={https://arxiv.org/abs/2602.11337}, 
}

@misc{kim2026cosmospolicyfinetuningvideo,
      title={Cosmos Policy: Fine-Tuning Video Models for Visuomotor Control and Planning}, 
      author={Moo Jin Kim and Yihuai Gao and Tsung-Yi Lin and Yen-Chen Lin and Yunhao Ge and Grace Lam and Percy Liang and Shuran Song and Ming-Yu Liu and Chelsea Finn and Jinwei Gu},
      year={2026},
      eprint={2601.16163},
      archivePrefix={arXiv},
      primaryClass={cs.AI},
      url={https://arxiv.org/abs/2601.16163}, 
}

@misc{deshpande2026molmob0tlargescalesimulationenables,
      title={MolmoB0T: Large-Scale Simulation Enables Zero-Shot Manipulation}, 
      author={Abhay Deshpande and Maya Guru and Rose Hendrix and Snehal Jauhri and Ainaz Eftekhar and Rohun Tripathi and Max Argus and Jordi Salvador and Haoquan Fang and Matthew Wallingford and Wilbert Pumacay and Yejin Kim and Quinn Pfeifer and Ying-Chun Lee and Piper Wolters and Omar Rayyan and Mingtong Zhang and Jiafei Duan and Karen Farley and Winson Han and Eli Vanderbilt and Dieter Fox and Ali Farhadi and Georgia Chalvatzaki and Dhruv Shah and Ranjay Krishna},
      year={2026},
      eprint={2603.16861},
      archivePrefix={arXiv},
      primaryClass={cs.RO},
      url={https://arxiv.org/abs/2603.16861}, 
}

@misc{qwenteam2026qwen35omnitechnicalreport,
      title={Qwen3.5-Omni Technical Report}, 
      author={Qwen Team},
      year={2026},
      eprint={2604.15804},
      archivePrefix={arXiv},
      primaryClass={cs.CL},
      url={https://arxiv.org/abs/2604.15804}, 
}

@misc{wu2023unleashinglargescalevideogenerative,
      title={Unleashing Large-Scale Video Generative Pre-training for Visual Robot Manipulation}, 
      author={Hongtao Wu and Ya Jing and Chilam Cheang and Guangzeng Chen and Jiafeng Xu and Xinghang Li and Minghuan Liu and Hang Li and Tao Kong},
      year={2023},
      eprint={2312.13139},
      archivePrefix={arXiv},
      primaryClass={cs.RO},
      url={https://arxiv.org/abs/2312.13139}, 
}

@misc{nvidia2025gr00tn1openfoundation,
      title={GR00T N1: An Open Foundation Model for Generalist Humanoid Robots}, 
      author={NVIDIA and : and Johan Bjorck and Fernando Castañeda and Nikita Cherniadev and Xingye Da and Runyu Ding and Linxi "Jim" Fan and Yu Fang and Dieter Fox and Fengyuan Hu and Spencer Huang and Joel Jang and Zhenyu Jiang and Jan Kautz and Kaushil Kundalia and Lawrence Lao and Zhiqi Li and Zongyu Lin and Kevin Lin and Guilin Liu and Edith Llontop and Loic Magne and Ajay Mandlekar and Avnish Narayan and Soroush Nasiriany and Scott Reed and You Liang Tan and Guanzhi Wang and Zu Wang and Jing Wang and Qi Wang and Jiannan Xiang and Yuqi Xie and Yinzhen Xu and Zhenjia Xu and Seonghyeon Ye and Zhiding Yu and Ao Zhang and Hao Zhang and Yizhou Zhao and Ruijie Zheng and Yuke Zhu},
      year={2025},
      eprint={2503.14734},
      archivePrefix={arXiv},
      primaryClass={cs.RO},
      url={https://arxiv.org/abs/2503.14734}, 
}

@misc{nvidia2025cosmosreason1physicalcommonsense,
      title={Cosmos-Reason1: From Physical Common Sense To Embodied Reasoning}, 
      author={NVIDIA and : and Alisson Azzolini and Junjie Bai and Hannah Brandon and Jiaxin Cao and Prithvijit Chattopadhyay and Huayu Chen and Jinju Chu and Yin Cui and Jenna Diamond and Yifan Ding and Liang Feng and Francesco Ferroni and Rama Govindaraju and Jinwei Gu and Siddharth Gururani and Imad El Hanafi and Zekun Hao and Jacob Huffman and Jingyi Jin and Brendan Johnson and Rizwan Khan and George Kurian and Elena Lantz and Nayeon Lee and Zhaoshuo Li and Xuan Li and Maosheng Liao and Tsung-Yi Lin and Yen-Chen Lin and Ming-Yu Liu and Xiangyu Lu and Alice Luo and Andrew Mathau and Yun Ni and Lindsey Pavao and Wei Ping and David W. Romero and Misha Smelyanskiy and Shuran Song and Lyne Tchapmi and Andrew Z. Wang and Boxin Wang and Haoxiang Wang and Fangyin Wei and Jiashu Xu and Yao Xu and Dinghao Yang and Xiaodong Yang and Zhuolin Yang and Jingxu Zhang and Xiaohui Zeng and Zhe Zhang},
      year={2025},
      eprint={2503.15558},
      archivePrefix={arXiv},
      primaryClass={cs.AI},
      url={https://arxiv.org/abs/2503.15558}, 
}
